\documentclass{article}

\usepackage[margin=1in]{geometry}
\usepackage{times}
\usepackage[round,authoryear]{natbib}

\usepackage[hidelinks]{hyperref}
\usepackage{etoc}
\usepackage{url}
\usepackage{latexsym}
\usepackage{amsmath,amssymb}
\usepackage{graphicx}
\usepackage{booktabs}
\usepackage{enumitem}
\usepackage{multirow}
\usepackage{float}
\usepackage{placeins}
\usepackage[most]{tcolorbox}
\usepackage[T1]{fontenc}
\usepackage[utf8]{inputenc}
\usepackage{microtype}
\usepackage{inconsolata}
\usepackage{dblfloatfix}

\definecolor{rfNativeBack}{RGB}{247,251,255}
\definecolor{rfNativeFrame}{RGB}{84,124,168}
\definecolor{rfNativeTitle}{RGB}{225,238,251}
\definecolor{rfAgentBack}{RGB}{247,251,249}
\definecolor{rfAgentFrame}{RGB}{58,119,104}
\definecolor{rfAgentTitle}{RGB}{226,242,235}
\hypersetup{
  pdftitle={ForeSci: Evaluating LLM Agents for Forward-Looking AI Research Judgment},
  pdfauthor={Qiuyu Tian, Haojie Yin, Yingce Xia, Youyong Kong, Zequn Liu, Hang Su, Jianghan Chao, Xiaowen Gu}
}
\title{ForeSci: Evaluating LLM Agents for Forward-Looking AI Research Judgment}
\date{}
\author{%
\begin{tabular}{c}
\textbf{Qiuyu Tian}$^{1,2}$ \quad
\textbf{Haojie Yin}$^{3}$ \quad
\textbf{Yingce Xia}$^{2}$ \quad
\textbf{Youyong Kong}$^{1}$ \\
\textbf{Zequn Liu}$^{2}$ \quad
\textbf{Hang Su}$^{2,4}$ \quad
\textbf{Jianghan Chao}$^{2,5}$ \quad
\textbf{Xiaowen Gu}$^{6,7}$ \\
\small $^{1}$Southeast University, Nanjing, China \\
\small $^{2}$Beijing Zhongguancun Academy, Beijing, China \\
\small $^{3}$Duke Kunshan University, Kunshan, China \\
\small $^{4}$East China Normal University \\
\small $^{5}$Gaoling School of Artificial Intelligence, Renmin University of China \\
\small $^{6}$School of Animation and Digital Arts, Communication University of China \\
\small $^{7}$Jiangsu Second Normal University
\end{tabular}%
}

\begin{document}

\maketitle

\begin{abstract}
AI research often requires decisions before future evidence exists: which bottleneck to attack,
which direction to pursue, or where a project should be positioned. We introduce ForeSci, a
temporally controlled benchmark for evaluating whether LLM agents can make such forward-looking
research judgements from historical evidence. ForeSci contains 500 tasks across four fast-moving AI
domains and four decision families. Each task is paired with a cutoff-aligned offline knowledge
base; post-cutoff papers are hidden during generation and used only for validation. To avoid random
future-event prediction, tasks are derived from pre-cutoff taxonomy branches and evidence signals,
while generation-time retrieval and tools are restricted to the cutoff-aligned evidence package. We evaluate native LLMs,
Hybrid RAG, and three research-agent adaptations across four backbones. Agent-based methods improve
traceability over Hybrid RAG, while their gains in future-target alignment over native LLMs are
modest and task dependent. Diagnostics reveal a recurring evidence-decision decoupling: agents may cite
relevant evidence while forecasting the wrong research object. Task-specific checks between evidence
and final decision reduce drift and improve future-target alignment. As research agents move from
assisting known tasks to shaping open research directions, ForeSci offers a first controlled step
toward measuring whether their judgement can be trusted to guide that future.
\end{abstract}

\etocdepthtag.toc{main}

\section{Introduction}

AI research moves on a timescale where today's frontier becomes tomorrow's
baseline. AI research decisions ({\em e.g.}, which bottleneck to attack or which direction
merits a six-month commitment) therefore require foresight. Decisions that align with the field's
subsequent trajectory can direct effort toward problems of lasting importance, whereas those that
do not risk committing substantial resources to priorities that quickly lose relevance. As
autonomous research agents become more widely deployed for ideation, planning, and scientific workflow execution
\citep{lu2026towards,li2024chainideas,tang2025airesearcher,yamada2025aiscientistv2,gridach2025agentic,chen2025mlrbench,lupidi2026airsbench,wang2025paperarena},
they are increasingly being asked to participate in this decision
layer. Whether current LLM agents
can make defensible, evidence-grounded research judgements that align with future developments yet
to unfold is therefore a central open question.

Existing benchmarks do not fully answer this question. Prior work has mostly evaluated whether AI systems can answer questions over papers, synthesize literature
\citep{lala2023paperqa,wan2024sciqag,lewis2020rag}, use tools
\citep{yao2023react,schick2023toolformer}, execute research workflows
\citep{chen2025mlrbench,lupidi2026airsbench,wang2025paperarena}, or generate predictions on specific future scientific facts
\citep{ajith2026prescience}. None of these tasks asks whether an agent can produce an open-ended
research decision, such as picking a bottleneck, ranking a research agenda,
or selecting a venue, using only the evidence available at a specific
historical moment.

Building such a benchmark raises two challenges. First, the evidence boundary must be enforceable. Post-cutoff papers should not appear in training corpus, retrieval or other model-visible benchmark artifacts. Otherwise, a system may rely on hindsight rather than foresight
\citep{zhao2024setclock,ye2024mirai,liu2025exante,ajith2026prescience,wang2026futurealigned}. Second, the research judgements being evaluated must be historically inferable. They should be grounded in signals available before the cutoff, rather than in arbitrary future events or design choices. A foresight benchmark must therefore govern both what a system can see and what it is fair to ask.

\begin{figure*}[t]
\centering
\includegraphics[width=\textwidth]{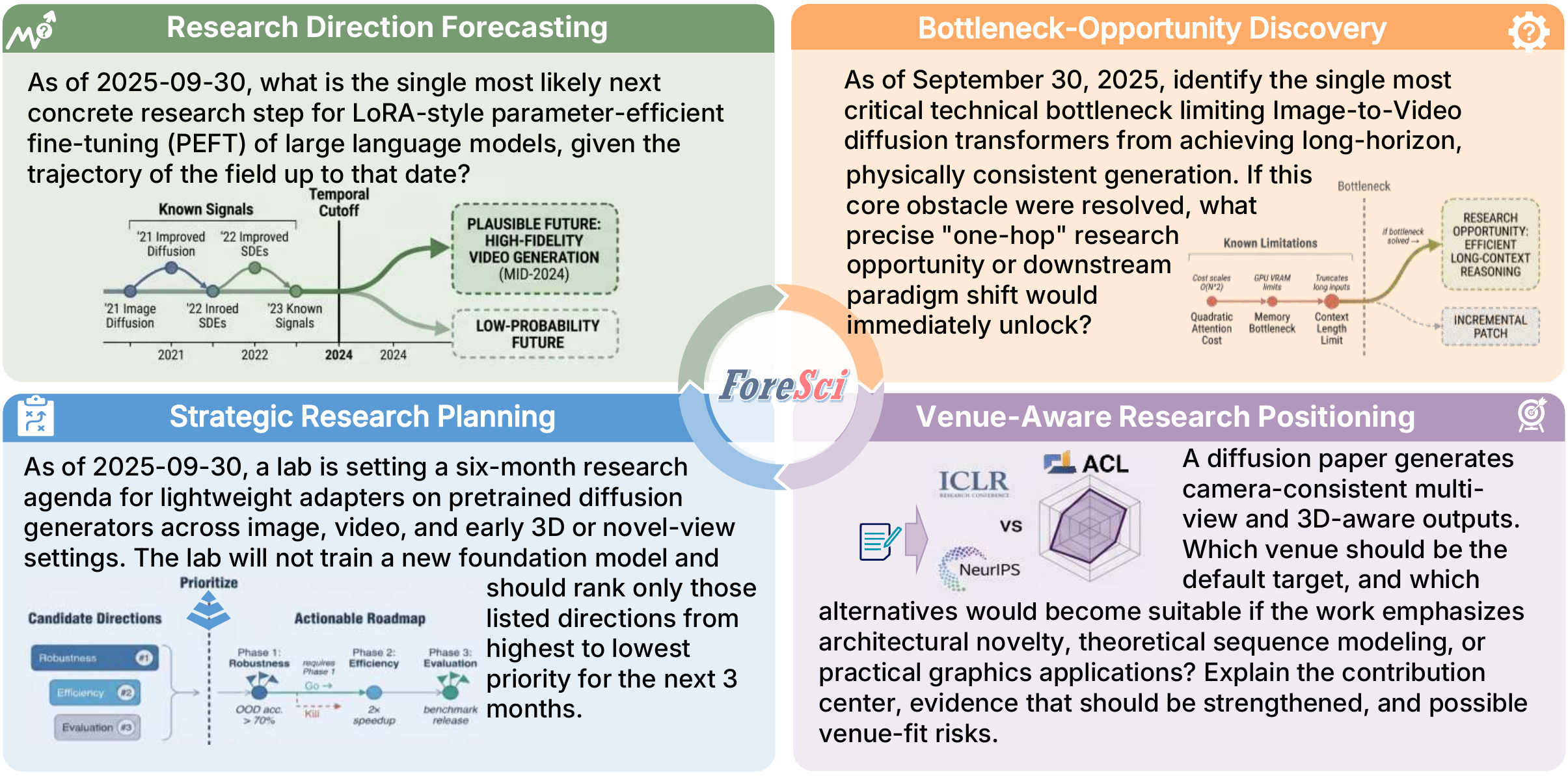}
\caption{Representative ForeSci task examples across the four decision families:
direction forecasting, bottleneck--opportunity discovery, strategic research planning, and
venue-aware research positioning.}
\label{fig:task-families}
\end{figure*}

To address these challenges, we introduce ForeSci, a temporally controlled benchmark for
forward-looking AI research judgement. It contains 500 tasks across four fast-moving AI domains and
four decision families (Figure \ref{fig:task-families}). Each task pairs a public question with a
cutoff-aligned offline knowledge base, while post-cutoff evidence is hidden until evaluation. Tasks are constructed from
pre-cutoff taxonomy branches, node-level evidence records, and method-evolution signals, ensuring
that each decision is historically inferable but not directly answerable from future leakage. Each answer is evaluated by four complementary signals: factual support following the atomic-fact~\citep{min2023factscore}, future-target
alignment~\citep{wang2026futurealigned}, evidence traceability, and reviewer persuasiveness motivated by peer-review reliability analyses~\citep{francois2015peerreview}. We evaluate a native LLM, Hybrid RAG, and three offline-adapted research-agent systems across four
LLM backbones. All systems operate within the same historical knowledge base, with web access disabled; we separately audit proprietary-backbone outputs for explicit post-cutoff target leakage.

Across the overall evaluation, agentic systems make the evidence chain more traceable, but do not consistently produce decisions that better match later research developments than native LLMs. The diagnostic audit identifies an evidence-decision decoupling: relevant evidence can still be connected to the wrong object, causal role, or intervention. Task-specific checks at the evidence-to-decision stage reduce drift and improve future-target alignment. Our key contributions include:
\begin{itemize}[leftmargin=*,noitemsep]
    \item \textbf{A temporally-controlled benchmark} with 500 tasks across four AI domains and four decision families, paired with cutoff-aligned offline knowledge bases and output-level temporal-leakage auditing
    \item \textbf{A multi-signal evaluation protocol} separating factuality, future-target alignment, evidence traceability, and reviewer persuasiveness.
    \item \textbf{A systematic evaluation and diagnostic audit of LLM research agents} showing task-conditional changes in traceability and factuality, and identifying a previously-unstudied failure mode—evidence-decision decoupling.
\end{itemize}

\section{Related Work}

\paragraph{Autonomous Research Agents}
AI-for-science systems have moved from local literature QA toward agentic workflows that retrieve,
synthesize, ideate, and execute parts of the research loop \cite{lu2026towards,ghareeb2026multi}. PaperQA-style systems
\citep{lala2023paperqa}, Chain-of-Ideas \citep{li2024chainideas}, AI-Researcher
\citep{tang2025airesearcher}, AI Scientist \citep{yamada2025aiscientistv2}, Intern-Atlas \cite{wu2026internatlas} and recent
agentic AI-for-science workflows \citep{gridach2025agentic} illustrate this shift toward autonomous
research assistance. As these agents are increasingly deployed for ideation and planning, they are implicitly asked to make research decisions. Yet whether they can do so from evidence available at a specific historical moment remains an open question. ForeSci directly evaluates this decision-making layer using cutoff-aligned historical evidence.

\paragraph{Benchmarks for Autonomous Research}
Existing benchmarks for autonomous research mainly focus on scientific reasoning \cite{lu2022scienceqa,center2026hle,bragg2026astabench,liu2025researchbench,jansen2025matteroffact}, artifacts—literature-grounded question answering \citep{wan2024sciqag, lala2023paperqa}, machine-learning research workflows \citep{chen2025mlrbench, lupidi2026airsbench}, and paper-based agent arenas \citep{wang2025paperarena}. These benchmarks measure retrieval, tool use, synthesis, or execution. Compared to them, ForeSci instead asks systems to make prospective research decisions rather than recover accessible answers or execute known workflows.
The benchmarks most closely related to our work aim to predict specific future scientific facts or artifacts. PreScience forecasts future scientific contributions and paper components \citep{ajith2026prescience}, while related efforts target future scientific events and progress, downstream insights, experimental outcomes, or emerging research topics \citep{wu2026forecasting,heyueya2026giants,sehwag2026scipredict,gu2024forecasting}. ForeSci moves beyond outcome-centric forecasting by testing whether agents can turn historical evidence into task-constrained decisions about what research to pursue next.

\paragraph{Temporal Integrity in Evaluation}
Temporal integrity is essential when evaluating foresight: without a strict cutoff, systems can
benefit from hindsight, leakage, or later-stabilized terminology rather than inference.
ExAnte~\citep{liu2025exante}, Set the Clock~\citep{zhao2024setclock}, ForecastBench~\citep{karger2025forecastbench}, FutureX \cite{zeng2025futurex}, FOReCAst~\citep{yuan2025forecast}, PROPHET~\citep{tao2025prophet}, and
MIRAI~\citep{ye2024mirai} all motivate time-sliced
evaluation for future-oriented reasoning.
Test of Time isolates temporal reasoning in synthetic settings \citep{fatemi2024testoftime}, while
AntiLeakBench constructs questions from updated real-world knowledge to reduce contamination
\citep{wu2025antileakbench}.
A line of work most closely related to ours uses future signals to evaluate generated research ideas, either by measuring their similarity to particular future papers \citep{wang2026futurealigned} or by using future-impact signals \citep{jiang2026hindsight,zhu2026sciimpact,ye2026proofoftime}. Compared to these approaches, ForeSci considers broader research judgment beyond a single idea, uses taxonomy-derived targets with human checks to separate historically inferable developments from arbitrary future events, and analyzes evidence-to-decision drift to explain how agents fail to connect available evidence to the appropriate research decision for the task at hand.

\section{The ForeSci Framework}
To systematically evaluate \emph{forward-looking AI research judgement}, ForeSci simulates a retrospective forecasting environment. Models are tasked with making research decisions at a strict historical cutoff, utilizing only chronologically aligned evidence.
\subsection{Problem Formulation}
Let $t$ denote a cutoff
date, $\mathcal{K}_{\leq t}(q)$ denote the cutoff-aligned knowledge base constructed for question $q$ (i.e., literature published up to $t$), and
$\mathcal{G}_{>t}(q)$ denote the withheld validation targets derived from post-cutoff literature. A benchmark
instance is
\begin{equation}
x=(q,t,\mathcal{K}_{\leq t}(q),f),
\end{equation}
where $f$ is the required task family. A system returns
$a=\pi_{\theta}(q,\mathcal{K}_{\leq t}(q))$ using only the provided cutoff-aligned knowledge base; $\mathcal{G}_{>t}(q)$ is accessible only to evaluation. To control model-visible information, we disable web search and allow systems to use only
$\mathcal{K}_{\leq t}(q)$ as external support when producing answers.

ForeSci instantiates this judgement problem through four task families:
\textit{Direction Forecasting}, \textit{Bottleneck--Opportunity Discovery},
\textit{Strategic Research Planning}, and \textit{Venue-Conditioned Positioning}. Each family asks
for a different research decision after $t$: predicting a concrete technical trajectory, identifying
a bottleneck and the opportunity it unlocks, ranking candidate research directions under planning
constraints, or positioning a project for an appropriate venue community.

\subsection{Data Collection and Filtering}
\label{sec:data-collection}

\begin{figure*}[t]
\centering
\includegraphics[width=\textwidth]{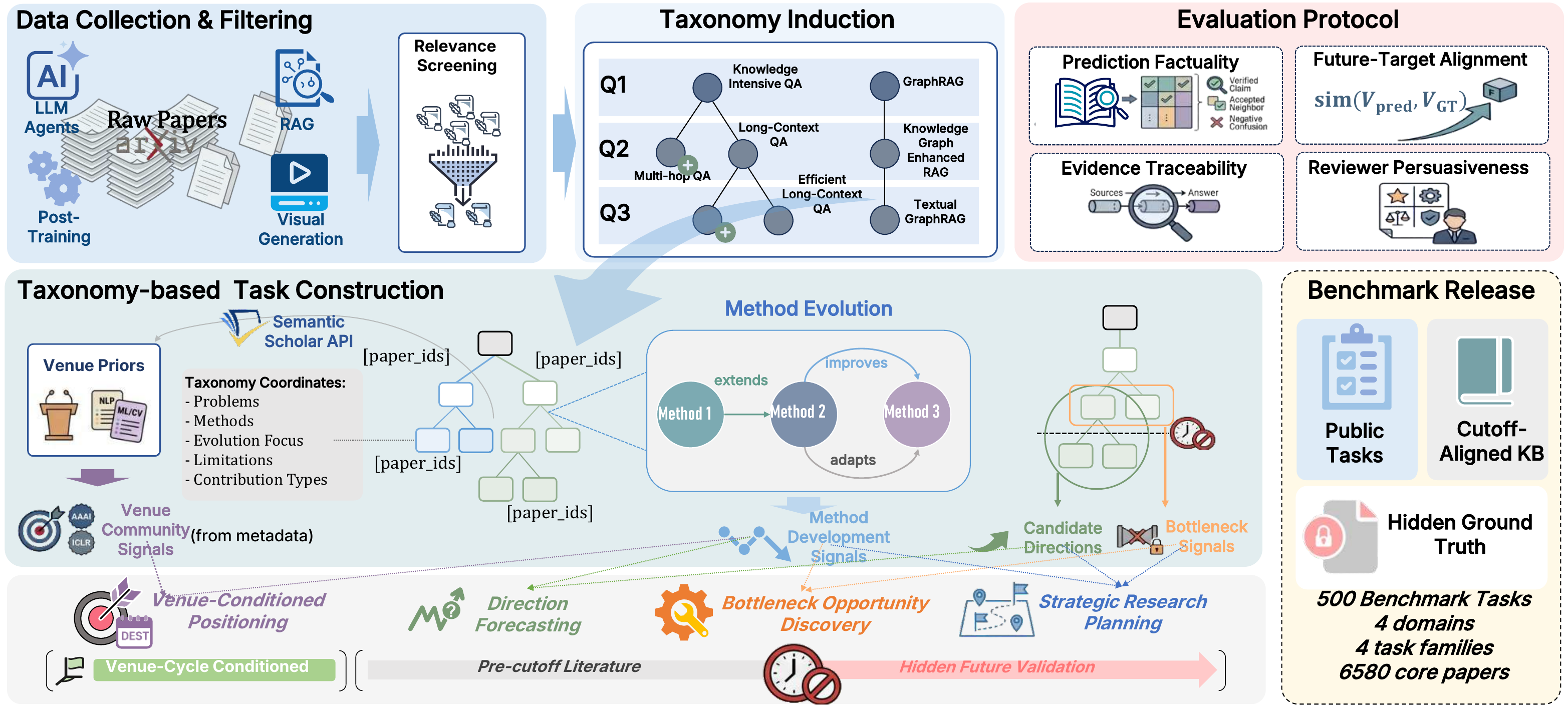}
\caption{Construction process for the current formal ForeSci release. The figure shows
the pipeline from corpus harvest and screening to temporal taxonomy induction, evidence and
evolution asset construction, task-family builders, hidden validation targets, and the final
benchmark release with public tasks and a paper knowledge base.}
\label{fig:benchmark-construction}
\end{figure*}

Figure~\ref{fig:benchmark-construction} summarizes the construction pipeline. ForeSci is
built from four rapidly evolving AI research areas: LLM agents, LLM fine-tuning and post-training,
RAG and retrieval structuring, and visual generative modeling. 
For each area, we harvest candidate papers from arXiv \footnote{\url{https://arxiv.org/}} using domain-specific queries, enrich
publication metadata with Semantic Scholar \footnote{\url{https://www.semanticscholar.org/}}, deduplicate arXiv identifiers, and retain
core/support papers after relevance and benchmark-core screening.

We apply two filtering stages to construct cutoff-aligned corpora. First, a domain-relevance screen removes
papers that only match surface keywords. Second, a stricter benchmark-core screen identifies representative papers
with central domain contributions and future-facing signals ({\em e.g.}, novel evaluation protocols, identified bottlenecks). Relevant but less central papers are retained as support papers,
noisy or borderline cases are excluded. Finally, the processed corpus is chronologically truncated at the cutoff time $t$ to form the public
pre-cutoff knowledge base $\mathcal{K}_{\leq t}$. The specific cutoff date $t$ varies across task instances, encompassing three-month (December 31, 2025), six-month (September 30, 2025), and venue-specific deadline settings after September 30, 2025. Domain and horizon statistics are reported in
Table~\ref{tab:domain-release-stats}, while monthly paper-count statistics are provided in
Figure~\ref{fig:monthly-raw-paper-counts}.
Additional construction details are provided in Appendix~\ref{sec:appendix-benchmark-artifacts}.

\begin{table}[t]
\centering
\small
\setlength{\tabcolsep}{5pt}
\caption{Domain and horizon statistics in ForeSci. KB document counts come from the cutoff-aligned
offline knowledge base. Three-month and six-month counts follow task-level horizon metadata;
venue-conditioned tasks use venue-cycle timing because their advisory and submission windows are
venue specific.}
\label{tab:domain-release-stats}
\begin{tabular}{@{}lrrrrr@{}}
\toprule
Domain & KB Docs & 3-month & 6-month & Venue cycle & Total tasks \\
\midrule
LLM Agents & 2,769 & 25 & 90 & 23 & 138 \\
LLM Fine-tuning and Post-training & 2,131 & 15 & 69 & 15 & 99 \\
RAG and Retrieval Structuring & 767 & 10 & 69 & 13 & 92 \\
Visual Generative Modeling and Diffusion & 913 & 19 & 78 & 74 & 171 \\
\midrule
Total & 6,580 & 69 & 306 & 125 & 500 \\
\bottomrule
\end{tabular}
\end{table}

\subsection{Taxonomy Construction}

To make the foresight problem both inferable and traceable, ForeSci models the evolution of AI research through taxonomy induction. This allows us to find specific research subdirections whose trajectories can be systematically deduced along the taxonomy and strictly grounded in historical evidence. 
We build on TaxoAdapt~\citep{kargupta2025taxoadapt} to induce this taxonomy as a graph representation of the evolving research landscape. For each domain $d$ and cutoff $t$, we induce a temporal taxonomy
\begin{equation}
\mathcal{T}_{d,t}=(\mathcal{V}_{d,t},\mathcal{E}_{d,t}),
\end{equation}
where nodes represent research subdirections and edges represent method-evolution relations (see Figure~\ref{fig:taxonomy-examples} for
illustrative examples). The taxonomy is dynamically expanded across sequential time slices of the cutoff-aligned corpus, preserving  temporal causality to prevent future information leakage.

\paragraph{Node representation.} Each node $v \in \mathcal{V}_{d,t}$ is aggregated from multiple cutoff-visible papers.
For each node $v$, we construct a \emph{node evidence record} that links the subdirection
back to the cutoff-visible literature. The record mainly includes the representative papers and supporting papers. Each paper has a \emph{full-text evidence} showing what problems, methods, evaluation focus, limitations, and contribution types had already appeared before $t$.

\paragraph{High-order signal extraction.} From these node evidence records, we derive high-order signals for downstream task construction:

(1) \emph{candidate directions}, which group one or more related nodes into coherent research options;

(2) \emph{method-development signals}~\citep{wu2026internatlas}, which record how methods, evaluations,
or bottlenecks evolve over time (see Figure \ref{fig:method-evolution} for an example);

(3) \emph{bottleneck signals}, which summarize recurring
limitations, evaluation gaps, reliability or safety concerns, dataset or benchmark needs, and
technical risks; 

(4) \emph{feasibility, dependency, and risk notes}, which record whether a candidate
direction is actionable as a near-term research plan; 

(5) \emph{venue-community metadata}, which summarize publication and community context, including contribution style,
maturity expectations, reviewer risks, and nearby venue contrasts.

All taxonomy structures are first extracted from cutoff-aligned evidence and then checked for
support strength and temporal validity. In an upstream audit, three annotators assessed 50 sampled
taxonomy edges and 50 sampled signals, yielding mean accuracies of 90.7\% for taxonomy induction
and 92.3\% for signal extraction. Every downstream task and target record is additionally reviewed
for temporal validity, leakage risk, specificity, support strength, and task clarity.

\subsection{Task Families}
\label{sec:task-families}

We derive four task families from the taxonomy. Task instances are constructed through a human--LLM collaborative
process: an LLM first drafts candidate questions, options, and answers from the
taxonomy-derived evidence records. Human experts then
inspect the source evidence, check cutoff validity and leakage risk, revise unclear or weakly
grounded items, and approve each final instance, ensuring that the benchmark instances reflect expert-validated scientific value and foresight challenges.

\paragraph{Direction Forecasting.}
This family asks the system to choose, from a fixed set of \emph{candidate directions}, which
direction is most likely to gain momentum in the post-cutoff window. The task $q$ is grounded in
\emph{node evidence records} and \emph{method-development signals} before $t$. 
The hidden future validation targets $\mathcal{G}_{>t}(q)$ are candidate directions (including primary directions and acceptable neighbors) with trajectory labels (e.g., \textsc{accelerating}, \textsc{steady}). They are initially induced by the \emph{method-development signals} after cutoff, combined with the necessary evidence before the cutoff. 

\paragraph{Bottleneck--Opportunity Discovery.}
This family asks the system to identify one root bottleneck in a cutoff-visible research
subdirection and explain what one-hop opportunity would open if that bottleneck were reduced. The
task $q$ is grounded in \emph{bottleneck signals}, \emph{full-text evidence}, and
\emph{method-development signals}. The hidden future validation
targets $\mathcal{G}_{>t}(q)$ are bottleneck--opportunity pairs, including primary bottlenecks,
acceptable bottleneck variants, unlocked opportunities, and mechanism descriptions. They are
initially induced from \emph{bottleneck signals} before $t$ and \emph{method-development signals}
after $t$, combined with the necessary evidence before the cutoff.

\paragraph{Strategic Research Planning.}
This family asks the system to rank a fixed set of research options for a hypothetical team making a
near-term research plan at the cutoff. The task $q$ is  derived from \emph{candidate directions} and
\emph{node evidence records}, \emph{method-development signals},
\emph{bottleneck signals}, \emph{feasibility, dependency,
and risk notes} before $t$. The hidden future validation targets $\mathcal{G}_{>t}(q)$ are ranked \emph{candidate directions}, including the preferred ordering, top-priority option, rationale units, milestones,
dependencies, risks, and go/no-go criteria. They are initially induced from post-cutoff \emph{method-development signals} and \emph{bottleneck signals}, combined with the necessary
evidence and \emph{feasibility, dependency,
and risk notes} before cutoff.

\paragraph{Venue-Conditioned Positioning.}
This family asks the system to position a proposed contribution for a target venue cycle.
Given a project description and five candidate venue families, the system selects one default
target and up to three alternatives, each with an explicit condition for suitability.
It also states the contribution center, three evidence upgrades, and two venue-fit risks.
Task construction draws on contribution types from \emph{full-text evidence} and \emph{venue-community metadata}.
The hidden future validation targets $\mathcal{G}_{>t}(q)$ are venue-positioning decisions. They are induced from
\emph{venue-community metadata} and
post-cutoff \emph{method-development signals} that reflect community expectations, combined with
the necessary evidence before the cutoff.

Across all families, public questions do not expose internal taxonomy information or post-cutoff
outcomes. The formal release contains 125 tasks for each family. Additional details on the benchmark construction are provided in
Appendix~\ref{sec:appendix-benchmark-artifacts}. For tasks with acceptable alternatives, the hidden targets retain multiple candidates or variants: all 125 Direction tasks include acceptable neighbors (mean 4.43 per task); 94/125 Bottleneck tasks include an acceptable bottleneck variant (mean 3.44 when nonempty), and 86/125 include an acceptable opportunity variant (mean 4.06 when nonempty); all 125 Venue tasks include acceptable venue-neighborhood entries (mean 4.10). Construction and scoring details given in Appendix~\ref{sec:appendix-benchmark-artifacts} and Appendix~\ref{sec:appendix-metric-details}.

\section{Evaluation}
\label{sec:evaluation}
\subsection{Metrics}
\label{sec:evaluation-metrics}

For each public question $q$, system answer $a$, pre-cutoff support packet
$\mathcal{E}_{\leq t}(q)$, and hidden future validation targets $\mathcal{G}_{>t}(q)$, we report
four complementary metrics. These metrics separately assess whether the answer states correct
future facts, reaches a conclusion consistent with the future target, grounds its reasoning in
visible pre-cutoff evidence, and presents a judgment persuasive to a reviewer.

\paragraph{Prediction Factuality (\emph{Fact}).}
This metric evaluates whether the answer makes claims that are supported by
$\mathcal{G}_{>t}(q)$. Following the atomic-fact view of FACTSCORE~\citep{min2023factscore}, we
extract atomic claims $\mathcal{C}(a)$ from the answer. We also define a hidden claim bank
$\mathcal{C}^{*}(q)\subset\mathcal{G}_{>t}(q)$: a set of task-relevant atomic validation claims
derived from hidden future validation targets. 
Prediction Factuality is their claim-level F1.

\paragraph{Future-Target Alignment (\emph{FTA}).}
This metric evaluates alignment with the family-specific future target in $\mathcal{G}_{>t}(q)$.
Direction and Bottleneck use reference-guided claim--target matching; Planning uses deterministic
ranking alignment. Venue uses a task-specific structured target because several venue communities
may be appropriate under different contribution framings. It scores canonical-family agreement,
conditional-alternative recall, and contribution-center alignment. Full definitions and the Venue
adaptation are given in Appendix~\ref{sec:appendix-metric-details}.

\paragraph{Evidence Traceability Score (\emph{Trace}).}
This metric measures whether the answer's key decisions are supported by pre-cutoff evidence.
For each task, we extract three fixed, family-specific decision slots from the final answer and
retrieve the three most similar evidence chunks per slot from the system's pre-cutoff support packet.
The evaluator assigns each slot a continuous score in $[0,1]$ based on the strength of evidence
support, the evidence-to-decision bridge, and unresolved conflict. Trace is the mean of the three
slot scores; Appendix~\ref{sec:appendix-metric-details} defines the slots and scoring criteria.
Answers without external support artifacts are reported as not applicable.

\paragraph{Reviewer Persuasiveness (\emph{Pers}).}
This metric estimates how persuasive an answer is to a reviewer making a research decision.
To score model-generated answers at scale, we train an XGBoost-based proxy for reviewer evaluation
\citep{chen2016xgboost}, following the component-based learned-evaluation approach of
USR and P2P-Eval \citep{mehri2020usr,sun2026p2p}. We select a balanced sample of 1,200 answers,
each independently rated by three annotators on a 1--7 rubric using only the public task and
anonymized final answer. We train a separate shallow regressor for each task family to predict
the mean rating from semantic content-quality and answer-form features, then normalize it to $[0,1]$.
Backbone, generation method, answer length, family identity, and hidden targets are excluded from $X$.
Five-fold task-grouped out-of-fold
validation yields Spearman $\rho=0.726$, MAE $0.551$ points, and within-task pairwise accuracy
$0.819$. The annotators show strong ordinal agreement (mean pairwise Spearman $\rho=0.818$ and
quadratic-weighted kappa $=0.821$). Appendix~\ref{sec:appendix-validation-analyses} provides the
full validation, and Tables~\ref{tab:pers-content-features}--\ref{tab:pers-form-features} list the
exact family-specific predictors. Figure~\ref{fig:pers-process} summarizes the complete annotation,
modeling, and scoring pipeline.

\subsection{Models, Systems, and Adaptation}
\label{sec:models-systems-adaptation}

We evaluate five systems: \textbf{Native LLM} without retrieval, \textbf{Hybrid RAG} with
sparse+dense retrieval, and three offline-adapted agentic systems: \textbf{CoI-style},
\textbf{ResearchAgent-style}, and \textbf{ARIS-style}. We adapt the agentic systems to ForeSci by constraining retrieval, tool use, and memory to
the offline knowledge base and by rendering final answers through task-family-specific output
schemas.  Detailed adaptation notes are in
Appendix~\ref{sec:appendix-agent-adaptations}. We evaluate Qwen3-235B~\citep{qwen2025qwen3}, GPT-5.2~\citep{openai2025gpt52model}, GLM-4.6~\citep{zai2025glm46}, and Gemini-3~\citep{google2025gemini3}. To prevent information leakage, we use only model versions with knowledge cutoff dates before September 30, 2025. All methods receive the same cutoff-aligned external evidence boundary.

\section{Results}

\subsection{Evaluation of LLM Agents}

\begin{table*}[t]
\centering
\scriptsize
\setlength{\tabcolsep}{3.2pt}
\renewcommand{\arraystretch}{1.05}
\caption{Overall results on ForeSci in terms of Prediction Factuality (Fact), Future-Target Alignment (FTA), Evidence Traceability Score (Trace), and Reviewer Persuasiveness (Pers). Bold marks the best method within the same backbone and metric. Table~\ref{tab:scalar-metric-stability} gives five-run stability views.
Table~\ref{tab:aggregate-method-contrasts} summarizes the aggregate method contrasts.}
\label{tab:task-averaged-metrics}
\resizebox{\textwidth}{!}{%
\begin{tabular}{@{}lrrrrrrrrrrrrrrrr@{}}
\toprule
& \multicolumn{4}{c}{Qwen3-235B} & \multicolumn{4}{c}{GPT-5.2} & \multicolumn{4}{c}{GLM-4.6} & \multicolumn{4}{c}{Gemini-3} \\
\cmidrule(lr){2-5}\cmidrule(lr){6-9}\cmidrule(lr){10-13}\cmidrule(l){14-17}
Method & Fact. & FTA & Trace & Pers & Fact. & FTA & Trace & Pers & Fact. & FTA & Trace & Pers & Fact. & FTA & Trace & Pers \\
\midrule
Native LLM
& 0.609 & 0.634 & -- & 0.591 & 0.615 & 0.619 & -- & 0.688 & 0.500 & 0.586 & -- & 0.503 & 0.559 & 0.627 & -- & 0.601 \\

Hybrid RAG
& 0.591 & 0.625 & 0.579 & 0.604 & 0.618 & 0.624 & 0.560 & 0.725 & 0.525 & 0.617 & 0.522 & 0.545 & 0.568 & 0.623 & 0.499 & 0.637 \\

CoI
& \textbf{0.612} & 0.637 & 0.580 & 0.640 & 0.632 & 0.640 & 0.648 & \textbf{0.748} & \textbf{0.555} & \textbf{0.657} & 0.538 & \textbf{0.569} & 0.575 & 0.624 & 0.546 & 0.642 \\

ResearchAgent
& 0.608 & \textbf{0.655} & 0.569 & 0.636 & \textbf{0.638} & 0.636 & \textbf{0.653} & 0.741 & 0.552 & 0.639 & 0.557 & 0.555 & \textbf{0.581} & \textbf{0.637} & 0.533 & \textbf{0.646} \\

ARIS
& 0.596 & 0.641 & \textbf{0.632} & \textbf{0.649} & 0.621 & \textbf{0.648} & 0.642 & 0.743 & 0.542 & 0.623 & \textbf{0.609} & 0.562 & 0.580 & 0.626 & \textbf{0.584} & 0.619 \\

\bottomrule
\end{tabular}
}
\end{table*}
\paragraph{Overall results}
Table~\ref{tab:task-averaged-metrics} reports Fact, FTA, Trace, and Pers across four backbones and
five methods.
On average, agentic methods improve evidence traceability over Hybrid RAG and reviewer
persuasiveness over native LLMs. Their gains over native LLMs on Fact and FTA, the two metrics
that directly assess alignment with future outcomes, are modest
(Table~\ref{tab:aggregate-method-contrasts}). These gains are also task-family dependent: no agentic
method is uniformly strongest across metrics, backbones, and task families, and in some settings,
agentic methods offer no clear advantage over native LLMs and even underperform Hybrid RAG
(Table~\ref{tab:metrics-split-by-family}). These findings reveal an important mismatch. Although
agentic methods can make greater use of pre-cutoff evidence and produce research-decision reports
that appear more persuasive to reviewers, the decisions themselves are not necessarily more
forward-looking. This mismatch creates a persuasive yet potentially misleading form of
pseudo-foresight. For example, a Qwen3-235B CoI-style planning answer achieves Trace = 0.892 but
FTA = 0.200 because it ranks embodied evaluation last due to immature tooling, whereas the
reference treats the same evaluation gap as the top-priority bottleneck
(Figure~\ref{fig:high-trace-high-drift-case}). Research decisions generated by agents therefore warrant particular scrutiny,
and this failure mode necessitates a closer analysis of why agentic systems produce such outputs.

\paragraph{Output-level temporal-leakage audit}
Because the knowledge boundaries of proprietary model versions are based primarily on provider
statements rather than transparently inspectable checkpoints, we additionally audit outputs from
GPT-5.2 and Gemini-3 for output-level temporal leakage. Specifically, we examine all 5,000 outputs
from these two proprietary backbones for explicit post-cutoff named entities. The initial screen
identifies 57 candidate mentions in 56 GPT-5.2 answers and 60 candidate mentions in 57 Gemini-3
answers. Manual adjudication confirms that none constitutes leakage; instead, the apparent overlaps
correspond to prospective proposals, providing further support for the effectiveness of our temporal
controls (Appendix~\ref{sec:appendix-output-leakage-audit} and
Table~\ref{tab:output-leakage-audit}).

\begin{figure*}[t]
    \centering
    \setlength{\parskip}{0pt}
    \setlength{\abovecaptionskip}{15pt}
    \includegraphics[width=\textwidth]{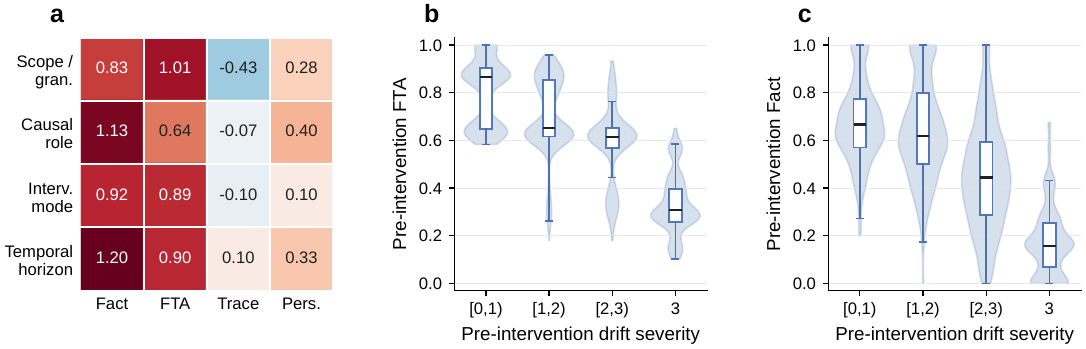}
    \par\nointerlineskip
    \caption{Evidence-to-decision drift and metric degradation.
    (a) Normalized metric drops associated with four drift dimensions; positive values indicate
    lower scores under moderate-to-severe drift.
    (b,c) FTA and Fact distributions across four mean-drift severity bins in the held-out sample.}
    \label{fig:drift-diagnostic-summary}
\end{figure*}

\subsection{Analysis and Intervention of Evidence-to-Decision Drift}
\label{sec:error-analysis}
\label{sec:intervention-mechanisms}

To understand the mismatch between traceability and future-facing Fact/FTA performance, we examine
where evidence-grounded answers diverge from future-aligned research decisions and whether targeted
controls can reduce that divergence.

\paragraph{Drift discovery and metric coupling}
Using a blinded 20-task discovery split for each family, we compare model answers with reference
decisions across all three agent-based methods and four backbones. This comparison reveals four recurring forms
of \emph{evidence-to-decision drift}: \emph{scope/granularity drift} selects the wrong level of
specificity; \emph{causal-role drift} confuses a cause, symptom, or opportunity;
\emph{intervention-mode drift} recommends the wrong kind of action; and
\emph{temporal-horizon drift} targets the wrong maturity stage. Each dimension is scored from 0 to
3, with higher values indicating a larger departure from the reference decision. Full annotation details and examples
are provided in Appendix~\ref{sec:appendix-bias-metric-coupling}.

We quantify each drift--metric association as the normalized score drop
$\Delta_{\mathrm{norm}}(m)=(\mathbb{E}[m\mid s=0]-\mathbb{E}[m\mid s\geq2])/\mathrm{SD}(m)$,
where $m$ is the target metric and $s$ is drift severity.
The four drift dimensions are strongly coupled to the metrics that directly evaluate decision
quality, but much less consistently to surface credibility. Moderate-to-severe drift is associated
with substantial drops in both Fact and FTA across all four dimensions, whereas its relationships
with Trace and Pers are weak or mixed (Figure~\ref{fig:drift-diagnostic-summary}a). This helps
explain the mismatch between traceability and future-facing Fact/FTA performance: an answer can remain traceable or persuasive while assigning
the evidence to the wrong decision object, causal role, action, or time horizon. This pattern also
extends beyond the discovery split. For example, in the held-out Qwen3-235B confirmation sample,
mean Drift severity is strongly and monotonically associated with lower FTA and Fact. Spearman correlation is
$-0.611$ for FTA (95\% task-cluster bootstrap CI $[-0.658,-0.559]$) and $-0.561$ for Fact
($[-0.610,-0.509]$) (Figure~\ref{fig:drift-diagnostic-summary}b,c).

\paragraph{Task-conditioned controls}
To test whether decision checks reduce drift and improve FTA and Fact, we introduce an
intervention between the evidence and decision stages. Specifically, we add task-specific decision
checks before final answer rendering, with task-specific implementations. For Bottleneck, after
collecting evidence, the check requires the agent to distinguish causal roles (e.g, observed failures,
proximal mechanisms, and upstream causes) before identifying the bottleneck. For Direction, it
checks whether new evidence matches the forecast's conceptual scope and temporal horizon. For
Planning, it compares candidate studies at a common scope and retains only evidence that supports
the decisive next-action comparison. For Venue, it checks that positioning remains anchored in the
project's current contribution rather than a broader topical scope or proposed future work.
Agent-specific prompts, commit rules, and evaluation details are deferred to the appendix
(Appendix~\ref{sec:appendix-drift-mechanisms}).

\begin{figure*}[t]
    \centering
    \setlength{\parskip}{0pt}
    \setlength{\abovecaptionskip}{15pt}
    \includegraphics[width=\textwidth]{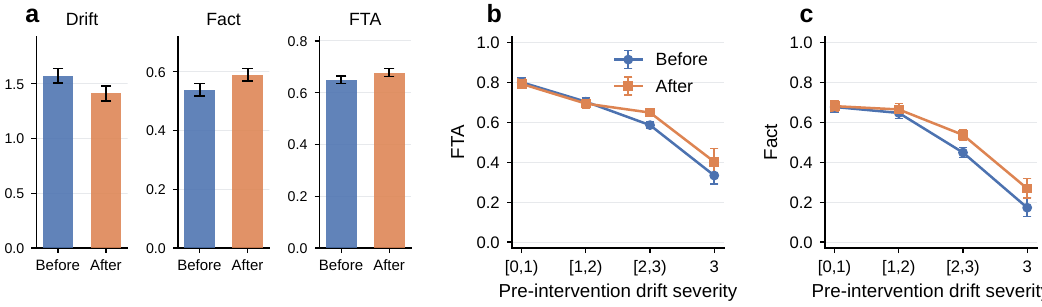}
    \par\nointerlineskip
    \caption{Evidence-to-decision intervention results.
    (a) Overall mean Drift, Fact, and FTA before and after intervention.
    (b,c) Mean FTA and Fact before and after intervention, stratified by pre-intervention Drift
    severity. Error bars denote 95\% task-cluster bootstrap confidence intervals.}
    \label{fig:drift-intervention-summary}
\end{figure*}

\paragraph{Intervention effects}
After intervention, mean Drift decreases by 0.161 (95\% task-cluster bootstrap CI
$[0.125,0.198]$), while Fact and FTA increase by 0.051 ($[0.038,0.064]$) and 0.028
($[0.015,0.040]$), respectively; all three intervals exclude zero
(Figure~\ref{fig:drift-intervention-summary}a).

The magnitude of these gains depends on initial Drift severity. For answers with low initial Drift,
FTA and Fact change only slightly. The separation between before and after becomes pronounced once
initial Drift reaches the moderate-to-severe range, with the largest Fact and FTA gains concentrated
in the two highest-severity bins (Figure~\ref{fig:drift-intervention-summary}b,c). This result suggests
a potential direction for future work: selectively gating research-agent revisions on
decision-level risk, so that additional controls focus on likely drift without unnecessarily
perturbing already aligned decisions.

\section{Conclusion}

ForeSci evaluates whether LLM agents can turn historically available evidence into forward-looking
AI research judgements. Its 500 cutoff-controlled tasks pair offline knowledge bases with hidden
post-cutoff validation targets across four decision families. Agentic workflows improve traceability
over Hybrid RAG and yield modest average gains in future-target alignment over native LLMs,
with substantial variation across tasks and backbones. The drift analysis identifies failures
in connecting evidence to the final decision, and task-specific controls improve alignment.
By separating factual
support, future-target alignment, traceability, and reviewer-style persuasiveness, ForeSci makes
these failures measurable.

\clearpage
\subsection*{AI use statement}

LLMs used as evaluated systems or benchmark evaluators are documented in the corresponding
methodology sections. Separately, generative AI assistants supported software prototyping and
debugging, experiment orchestration, preliminary result checking, language editing, translation,
LaTeX formatting, and manuscript drafting. The authors independently reviewed and verified all
AI-assisted code, analyses, results, and text. The benchmark methodology, data processing,
experimental procedures, and reported results were checked by the authors, who take responsibility
for the final manuscript, including its claims, analyses, and artifacts.

\subsection*{Reproducibility statement}

Section~3 provides the formal benchmark and task definitions, and Section~4 specifies the metrics,
model and system settings, and evaluation protocol. Appendix~\ref{sec:appendix-benchmark-artifacts}
documents benchmark construction, temporal controls, selection parameters, and artifact separation;
Appendix~\ref{sec:appendix-evaluation-details} gives metric calibration, human annotation, validation
analyses, and evaluation prompts; and Appendix~\ref{sec:appendix-agent-adaptations} describes the
agent adaptations and comparison controls. The release manifest in
Appendix~\ref{sec:appendix-responsible-research} enumerates the benchmark tasks, cutoff-aligned
knowledge bases, evaluation goldsets, prompts, schemas, scripts, run manifests, checksums, and
result-building artifacts. Code and supplementary materials will be released in anonymized form for
review and made publicly available upon publication.

\subsection*{Ethics statement}

ForeSci is constructed from publicly available scholarly artifacts and does not use private or
privacy-sensitive user data. Human-validation results are reported only in aggregate, and individual
annotator records are not released. The benchmark is intended to evaluate automated research
systems under controlled historical evidence, not to automate peer review, venue selection, hiring,
funding, or research-prioritization decisions. Its outputs are evaluation artifacts rather than
validated predictions or submission-ready manuscripts, and subsequent use requires human
verification and responsibility.

\begingroup
\makeatletter
\let\rfoldthebibliography\thebibliography
\renewcommand{\thebibliography}[1]{%
  \rfoldthebibliography{#1}%
  \setlength{\itemsep}{0.2ex}%
  \setlength{\parsep}{0pt}%
}
\makeatother
\bibliography{references}

\begin{thebibliography}{48}
\providecommand{\natexlab}[1]{#1}
\providecommand{\url}[1]{\texttt{#1}}
\expandafter\ifx\csname urlstyle\endcsname\relax
  \providecommand{\doi}[1]{doi: #1}\else
  \providecommand{\doi}{doi: \begingroup \urlstyle{rm}\Url}\fi

\bibitem[Ajith et~al.(2026)Ajith, Singh, DeYoung, Kunievsky, Kozlowski,
  Tafjord, Evans, Weld, Hope, and Downey]{ajith2026prescience}
Anirudh Ajith, Amanpreet Singh, Jay DeYoung, Nadav Kunievsky, Austin~C.
  Kozlowski, Oyvind Tafjord, James Evans, Daniel~S. Weld, Tom Hope, and Doug
  Downey.
\newblock Prescience: A benchmark for forecasting scientific contributions.
\newblock \emph{arXiv preprint arXiv:2602.20459}, 2026.
\newblock \doi{10.48550/arXiv.2602.20459}.
\newblock URL \url{https://arxiv.org/abs/2602.20459}.

\bibitem[Bragg et~al.(2026)Bragg, D'Arcy, Balepur, Bareket, Dalvi~Mishra,
  Feldman, Haddad, Hwang, Jansen, Kishore, Majumder, Naik, Rahamimov,
  Richardson, Singh, Surana, Tiktinsky, Vasu, Wiener, Anastasiades, Candra,
  Dunkelberger, Emery, Evans, Hamada, Huff, Kinney, Latzke, Lochner,
  Lozano-Aguilera, Nguyen, Rao, Tanaka, Vlahos, Clark, Downey, Goldberg,
  Sabharwal, and Weld]{bragg2026astabench}
Jonathan Bragg, Mike D'Arcy, Nishant Balepur, Dan Bareket, Bhavana
  Dalvi~Mishra, Sergey Feldman, Dany Haddad, Jena~D. Hwang, Peter Jansen,
  Varsha Kishore, Bodhisattwa~Prasad Majumder, Aakanksha Naik, Sigal Rahamimov,
  Kyle Richardson, Amanpreet Singh, Harshit Surana, Aryeh Tiktinsky, Rosni
  Vasu, Guy Wiener, Chloe Anastasiades, Stefanus Candra, Jason Dunkelberger,
  Daniel Emery, Rob Evans, Malachi Hamada, Regan Huff, Rodney Kinney, Matt
  Latzke, Jaron Lochner, Ruben Lozano-Aguilera, Ngoc-Uyen Nguyen, Smita Rao,
  Amber Tanaka, Brooke Vlahos, Peter Clark, Doug Downey, Yoav Goldberg, Ashish
  Sabharwal, and Daniel~S. Weld.
\newblock Astabench: Rigorous benchmarking of {AI} agents with a scientific
  research suite.
\newblock In \emph{International Conference on Learning Representations}, pages
  110136--110223, 2026.
\newblock URL
  \url{https://proceedings.iclr.cc/paper_files/paper/2026/hash/b2ce9568dbb559aefc8c98ca5b5314ce-Abstract-Conference.html}.

\bibitem[{Center for AI Safety} et~al.(2026){Center for AI Safety}, {Scale AI},
  and {HLE Contributors Consortium}]{center2026hle}
{Center for AI Safety}, {Scale AI}, and {HLE Contributors Consortium}.
\newblock A benchmark of expert-level academic questions to assess {AI}
  capabilities.
\newblock \emph{Nature}, 649:\penalty0 1139--1146, 2026.

\bibitem[Chen et~al.(2025)Chen, Xiong, Lu, Han, Deng, He, Wu, Li, Liu, and
  Hooi]{chen2025mlrbench}
Hui Chen, Miao Xiong, Yujie Lu, Wei Han, Ailin Deng, Yufei He, Jiaying Wu, Yibo
  Li, Yue Liu, and Bryan Hooi.
\newblock Mlr-bench: Evaluating ai agents on open-ended machine learning
  research.
\newblock \emph{Advances in Neural Information Processing Systems}, 38, 2025.

\bibitem[Chen and Guestrin(2016)]{chen2016xgboost}
Tianqi Chen and Carlos Guestrin.
\newblock {XGBoost}: A scalable tree boosting system.
\newblock In \emph{Proceedings of the 22nd ACM SIGKDD International Conference
  on Knowledge Discovery and Data Mining}, pages 785--794, 2016.
\newblock \doi{10.1145/2939672.2939785}.
\newblock URL \url{https://doi.org/10.1145/2939672.2939785}.

\bibitem[Fatemi et~al.(2025)Fatemi, Kazemi, Tsitsulin, Malkan, Yim, Palowitch,
  Seo, Halcrow, and Perozzi]{fatemi2024testoftime}
Bahare Fatemi, Seyed~Mehran Kazemi, Anton Tsitsulin, Karishma Malkan, Jinyeong
  Yim, John Palowitch, Sungyong Seo, Jonathan Halcrow, and Bryan Perozzi.
\newblock Test of time: A benchmark for evaluating {LLM}s on temporal
  reasoning.
\newblock In \emph{International Conference on Learning Representations}, pages
  94426--94447, 2025.
\newblock URL
  \url{https://proceedings.iclr.cc/paper_files/paper/2025/hash/eb7295a8bc613b375726659c2ecd6f14-Abstract-Conference.html}.

\bibitem[Francois(2015)]{francois2015peerreview}
Olivier Francois.
\newblock Arbitrariness of peer review: A bayesian analysis of the {NIPS}
  experiment.
\newblock \emph{arXiv preprint arXiv:1507.06411}, 2015.
\newblock \doi{10.48550/arXiv.1507.06411}.
\newblock URL \url{https://arxiv.org/abs/1507.06411}.

\bibitem[Ghareeb et~al.(2026)Ghareeb, Chang, Mitchener, Yiu, Szostkiewicz,
  Shved, Gyimesi, Laurent, Wright, Razzak, White, Finnemann, Hinks, and
  Rodriques]{ghareeb2026multi}
Ali~Essam Ghareeb, Benjamin Chang, Ludovico Mitchener, Angela Yiu, Caralyn~J.
  Szostkiewicz, Dmytro Shved, Gavin~J. Gyimesi, Jon~M. Laurent, Samantha~M.
  Wright, Muhammed~T. Razzak, Andrew~D. White, Silvia~C. Finnemann, Michaela~M.
  Hinks, and Samuel~G. Rodriques.
\newblock A multi-agent system for automating scientific discovery.
\newblock \emph{Nature}, 655\penalty0 (8122):\penalty0 497--505, 2026.
\newblock \doi{10.1038/s41586-026-10652-y}.
\newblock URL \url{https://www.nature.com/articles/s41586-026-10652-y}.

\bibitem[{Google}(2025)]{google2025gemini3}
{Google}.
\newblock Gemini models.
\newblock \url{https://ai.google.dev/gemini-api/docs/models}, 2025.
\newblock Accessed 2026-05-26.

\bibitem[Gridach et~al.(2025)Gridach, Nanavati, Mack, El~Abidine, and
  Mendes]{gridach2025agentic}
Mourad Gridach, Jay Nanavati, Christina Mack, Khaldoun~Zine El~Abidine, and
  Lenon Mendes.
\newblock Agentic {AI} for scientific discovery: A survey of progress,
  challenges, and future directions.
\newblock In \emph{ICLR 2025 Workshops: AgenticAI}, 2025.
\newblock URL
  \url{https://mlanthology.org/iclrw/2025/gridach2025iclrw-agentic/}.

\bibitem[Gu and Krenn(2025)]{gu2024forecasting}
Xuemei Gu and Mario Krenn.
\newblock Forecasting high-impact research topics via machine learning on
  evolving knowledge graphs.
\newblock \emph{Machine Learning: Science and Technology}, 6\penalty0
  (2):\penalty0 025041, 2025.
\newblock \doi{10.1088/2632-2153/add6ef}.
\newblock URL \url{https://doi.org/10.1088/2632-2153/add6ef}.

\bibitem[He-Yueya et~al.(2026)He-Yueya, Singh, Gao, Li, Yang, Finn, Brunskill,
  and Goodman]{heyueya2026giants}
Joy He-Yueya, Anikait Singh, Ge~Gao, Michael~Y. Li, Sherry Yang, Chelsea Finn,
  Emma Brunskill, and Noah~D. Goodman.
\newblock Giants: Generative insight anticipation from scientific literature.
\newblock \emph{arXiv preprint arXiv:2604.09793}, 2026.

\bibitem[Jansen et~al.(2025)Jansen, Hassan, and Wang]{jansen2025matteroffact}
Peter Jansen, Samiah Hassan, and Ruoyao Wang.
\newblock Matter-of-fact: A benchmark for verifying the feasibility of
  literature-supported claims in materials science.
\newblock In \emph{Empirical Methods in Natural Language Processing}, 2025.

\bibitem[Jiang(2026)]{jiang2026hindsight}
Bo~Jiang.
\newblock {HindSight}: Evaluating {LLM}-generated research ideas via future
  impact.
\newblock \emph{arXiv preprint arXiv:2603.15164}, 2026.
\newblock \doi{10.48550/arXiv.2603.15164}.
\newblock URL \url{https://arxiv.org/abs/2603.15164}.

\bibitem[Karger et~al.(2025)Karger, Bastani, Chen, Jacobs, Halawi, Zhang, and
  Tetlock]{karger2025forecastbench}
Ezra Karger, Houtan Bastani, Yueh-Han Chen, Zachary Jacobs, Danny Halawi, Fred
  Zhang, and Philip Tetlock.
\newblock Forecastbench: A dynamic benchmark of {AI} forecasting capabilities.
\newblock In \emph{International Conference on Learning Representations}, 2025.

\bibitem[Kargupta et~al.(2025)Kargupta, Zhang, Zhang, Zhang, Mitra, and
  Han]{kargupta2025taxoadapt}
Priyanka Kargupta, Nan Zhang, Yunyi Zhang, Rui Zhang, Prasenjit Mitra, and
  Jiawei Han.
\newblock Taxoadapt: Aligning llm-based multidimensional taxonomy construction
  to evolving research corpora.
\newblock In \emph{Proceedings of the 63rd Annual Meeting of the Association
  for Computational Linguistics (Volume 1: Long Papers)}, pages 29834--29850,
  2025.

\bibitem[L{\'a}la et~al.(2023)L{\'a}la, O'Donoghue, Shtedritski, Cox,
  Rodriques, and White]{lala2023paperqa}
Jakub L{\'a}la, Odhran O'Donoghue, Aleksandar Shtedritski, Sam Cox, Samuel~G.
  Rodriques, and Andrew~D. White.
\newblock Paperqa: Retrieval-augmented generative agent for scientific
  research.
\newblock \emph{arXiv preprint arXiv:2312.07559}, 2023.
\newblock \doi{10.48550/arXiv.2312.07559}.
\newblock URL \url{https://arxiv.org/abs/2312.07559}.

\bibitem[Lewis et~al.(2020)Lewis, Perez, Piktus, Petroni, Karpukhin, Goyal,
  Küttler, Lewis, Yih, Rockt{\"a}schel, Riedel, and Kiela]{lewis2020rag}
Patrick Lewis, Ethan Perez, Aleksandra Piktus, Fabio Petroni, Vladimir
  Karpukhin, Naman Goyal, Heinrich Küttler, Mike Lewis, Wen-tau Yih, Tim
  Rockt{\"a}schel, Sebastian Riedel, and Douwe Kiela.
\newblock Retrieval-augmented generation for knowledge-intensive nlp tasks.
\newblock In \emph{Advances in Neural Information Processing Systems},
  volume~33, pages 9459--9474. Curran Associates, Inc., 2020.
\newblock URL
  \url{https://proceedings.neurips.cc/paper/2020/hash/6b493230205f780e1bc26945df7481e5-Abstract.html}.

\bibitem[Li et~al.(2025)Li, Xu, Guo, Zhao, Li, Yuan, Zhang, Jiang, Xin, Dang,
  Rong, Zhao, Feng, and Bing]{li2024chainideas}
Long Li, Weiwen Xu, Jiayan Guo, Ruochen Zhao, Xingxuan Li, Yuqian Yuan, Boqiang
  Zhang, Yuming Jiang, Yifei Xin, Ronghao Dang, Yu~Rong, Deli Zhao, Tian Feng,
  and Lidong Bing.
\newblock Chain of ideas: Revolutionizing research via novel idea development
  with {LLM} agents.
\newblock In \emph{Findings of the Association for Computational Linguistics:
  EMNLP 2025}, pages 8971--9004. Association for Computational Linguistics,
  2025.
\newblock \doi{10.18653/v1/2025.findings-emnlp.477}.
\newblock URL \url{https://aclanthology.org/2025.findings-emnlp.477/}.

\bibitem[Liu et~al.(2026{\natexlab{a}})Liu, Wei, Shi, Li, Zhang, Dhillon, and
  Mei]{liu2025exante}
Yachuan Liu, Xiaochun Wei, Lin Shi, Xinnuo Li, Bohan Zhang, Paramveer~S
  Dhillon, and Qiaozhu Mei.
\newblock Exante: A benchmark for ex-ante inference in large language models.
\newblock In \emph{Proceedings of the 19th Conference of the European Chapter
  of the Association for Computational Linguistics (Volume 1: Long Papers)},
  pages 1551--1571, 2026{\natexlab{a}}.

\bibitem[Liu et~al.(2026{\natexlab{b}})Liu, Yang, Xie, Ni, Gao, Li, Tang,
  Ouyang, Cambria, and Zhou]{liu2025researchbench}
Yujie Liu, Zonglin Yang, Tong Xie, Jinjie Ni, Ben Gao, Yuqiang Li, Shixiang
  Tang, Wanli Ouyang, Erik Cambria, and Dongzhan Zhou.
\newblock {ResearchBench}: Benchmarking {LLM}s in scientific discovery via
  inspiration-based task decomposition.
\newblock In \emph{Findings of the Association for Computational Linguistics:
  ACL 2026}, pages 13187--13207. Association for Computational Linguistics,
  2026{\natexlab{b}}.
\newblock \doi{10.18653/v1/2026.findings-acl.644}.
\newblock URL \url{https://aclanthology.org/2026.findings-acl.644/}.

\bibitem[Lu et~al.(2026)Lu, Lu, Lange, Yamada, Hu, Foerster, Ha, and
  Clune]{lu2026towards}
Chris Lu, Cong Lu, Robert~Tjarko Lange, Yutaro Yamada, Shengran Hu, Jakob
  Foerster, David Ha, and Jeff Clune.
\newblock Towards end-to-end automation of ai research.
\newblock \emph{Nature}, 651\penalty0 (8107):\penalty0 914--919, 2026.

\bibitem[Lu et~al.(2022)Lu, Mishra, Xia, Qiu, Chang, Zhu, Tafjord, Clark, and
  Kalyan]{lu2022scienceqa}
Pan Lu, Swaroop Mishra, Tanglin Xia, Liang Qiu, Kai-Wei Chang, Song-Chun Zhu,
  Oyvind Tafjord, Peter Clark, and Ashwin Kalyan.
\newblock Learn to explain: Multimodal reasoning via thought chains for science
  question answering.
\newblock In \emph{Advances in Neural Information Processing Systems}, 2022.

\bibitem[Lupidi et~al.(2026)Lupidi, Gauri, Foster, Al~Omari, Magka, Pepe,
  Audran-Reiss, Aghamelu, Baldwin, Cipolina-Kun, Gagnon-Audet, Leow, Lefdal,
  Mossalam, Moudgil, Nazir, Tewolde, Urrego, Armengol~Estape, Budhiraja,
  Chaurasia, Charnalia, Dunfield, Hambardzumyan, Izcovich, Josifoski,
  Mediratta, Niu, Pathak, Shvartsman, Toledo, Protopopov, Raileanu, Miller,
  Shavrina, Foerster, and Bachrach]{lupidi2026airsbench}
Alisia Lupidi, Bhavul Gauri, Thomas~Simon Foster, Bassel Al~Omari, Despoina
  Magka, Alberto Pepe, Alexis Audran-Reiss, Muna Aghamelu, Nicolas Baldwin,
  Lucia Cipolina-Kun, Jean-Christophe Gagnon-Audet, Chee~Hau Leow, Sandra
  Lefdal, Hossam Mossalam, Abhinav Moudgil, Saba Nazir, Emanuel Tewolde, Isabel
  Urrego, Jordi Armengol~Estape, Amar Budhiraja, Gaurav Chaurasia, Abhishek
  Charnalia, Derek Dunfield, Karen Hambardzumyan, Daniel Izcovich, Martin
  Josifoski, Ishita Mediratta, Kelvin Niu, Parth Pathak, Michael Shvartsman,
  Edan Toledo, Anton Protopopov, Roberta Raileanu, Alexander Miller, Tatiana
  Shavrina, Jakob Foerster, and Yoram Bachrach.
\newblock {AIRS}-bench: a suite of tasks for frontier {AI} research science
  agents.
\newblock \emph{arXiv preprint arXiv:2602.06855}, 2026.
\newblock \doi{10.48550/arXiv.2602.06855}.
\newblock URL \url{https://arxiv.org/abs/2602.06855}.

\bibitem[Mehri and Eskenazi(2020)]{mehri2020usr}
Shikib Mehri and Maxine Eskenazi.
\newblock {USR}: An unsupervised and reference-free evaluation metric for
  dialog generation.
\newblock In \emph{Proceedings of the 58th Annual Meeting of the Association
  for Computational Linguistics}, pages 681--707, 2020.
\newblock \doi{10.18653/v1/2020.acl-main.64}.
\newblock URL \url{https://aclanthology.org/2020.acl-main.64/}.

\bibitem[Min et~al.(2023)Min, Krishna, Lyu, Lewis, Yih, Koh, Iyyer,
  Zettlemoyer, and Hajishirzi]{min2023factscore}
Sewon Min, Kalpesh Krishna, Xinxi Lyu, Mike Lewis, Wen-tau Yih, Pang Koh, Mohit
  Iyyer, Luke Zettlemoyer, and Hannaneh Hajishirzi.
\newblock Factscore: Fine-grained atomic evaluation of factual precision in
  long form text generation.
\newblock In \emph{Proceedings of the 2023 Conference on Empirical Methods in
  Natural Language Processing}, pages 12076--12100, 2023.

\bibitem[{OpenAI}(2025)]{openai2025gpt52model}
{OpenAI}.
\newblock {GPT-5.2}.
\newblock \url{https://openai.com/index/introducing-gpt-5-2/}, 2025.
\newblock Accessed 2026-05-26.

\bibitem[{Qwen Team}(2025)]{qwen2025qwen3}
{Qwen Team}.
\newblock {Qwen3}: Think deeper, act faster.
\newblock \url{https://qwenlm.github.io/blog/qwen3/}, April 2025.
\newblock Accessed 2026-05-26.

\bibitem[Schick et~al.(2023)Schick, Dwivedi-Yu, Dess{\`\i}, Raileanu, Lomeli,
  Hambro, Zettlemoyer, Cancedda, and Scialom]{schick2023toolformer}
Timo Schick, Jane Dwivedi-Yu, Roberto Dess{\`\i}, Roberta Raileanu, Maria
  Lomeli, Eric Hambro, Luke Zettlemoyer, Nicola Cancedda, and Thomas Scialom.
\newblock Toolformer: Language models can teach themselves to use tools.
\newblock \emph{Advances in neural information processing systems},
  36:\penalty0 68539--68551, 2023.

\bibitem[Sehwag et~al.(2026)Sehwag, Lau, Ehsani~Oskouie, Shabihi, Liang,
  Toledo, Mangialardi, Fonrouge, Hernandez~Cardona, Vergara, Tyagi, Zhang,
  Bhatter, Johnson, Huang, Hernandez~Montoya, and Liu]{sehwag2026scipredict}
Udari~Madhushani Sehwag, Elaine Lau, Haniyeh Ehsani~Oskouie, Shayan Shabihi,
  Erich Liang, Andrea Toledo, Guillermo Mangialardi, Sergio Fonrouge,
  Ed-Yeremai Hernandez~Cardona, Paula Vergara, Utkarsh Tyagi, Chen Bo~Calvin
  Zhang, Pavi Bhatter, Nicholas Johnson, Furong Huang, Ernesto~Gabriel
  Hernandez~Montoya, and Bing Liu.
\newblock Scipredict: Can llms predict the outcomes of scientific experiments
  in natural sciences?
\newblock \emph{arXiv preprint arXiv:2604.10718}, 2026.

\bibitem[Sun et~al.(2026)Sun, Pan, Yang, Sui, Shi, Cheng, Li, Zhang, Huang,
  Yang, and Li]{sun2026p2p}
Tao Sun, Enhao Pan, Zhengkai Yang, Kaixin Sui, Jiajun Shi, Xianfu Cheng,
  Tongliang Li, Ge~Zhang, Wenhao Huang, Jian Yang, and Zhoujun Li.
\newblock {P2P}: Automated paper-to-poster generation and fine-grained
  benchmark.
\newblock In \emph{International Conference on Learning Representations}, 2026.
\newblock URL
  \url{https://proceedings.iclr.cc/paper_files/paper/2026/hash/1fe6f635fe265292aba3987b5123ae3d-Abstract-Conference.html}.

\bibitem[Tang et~al.(2025)Tang, Xia, Li, and Huang]{tang2025airesearcher}
Jiabin Tang, Lianghao Xia, Zhonghang Li, and Chao Huang.
\newblock Ai-researcher: Autonomous scientific innovation.
\newblock \emph{Advances in Neural Information Processing Systems},
  38:\penalty0 9481--9520, 2025.

\bibitem[Tao et~al.(2025)Tao, Wu, Jin, Bai, Zhao, Dou, Chen, Li, Li, Tao, and
  Zhang]{tao2025prophet}
Zhengwei Tao, Pu~Wu, Zhi Jin, Xiaoying Bai, Haiyan Zhao, Chengfeng Dou, Xiancai
  Chen, Jia Li, Linyu Li, Chongyang Tao, and Wentao Zhang.
\newblock {PROPHET}: An inferable future forecasting benchmark with causal
  intervened likelihood estimation.
\newblock \emph{arXiv preprint arXiv:2504.01509}, 2025.

\bibitem[Wan et~al.(2024)Wan, Liu, Ajith, Grazian, Hoex, Zhang, Kit, Xie, and
  Foster]{wan2024sciqag}
Yuwei Wan, Yixuan Liu, Aswathy Ajith, Clara Grazian, Bram Hoex, Wenjie Zhang,
  Chunyu Kit, Tong Xie, and Ian Foster.
\newblock Sciqag: A framework for auto-generated science question answering
  dataset with fine-grained evaluation.
\newblock \emph{arXiv preprint arXiv:2405.09939}, 2024.
\newblock \doi{10.48550/arXiv.2405.09939}.
\newblock URL \url{https://arxiv.org/abs/2405.09939}.

\bibitem[Wang et~al.(2025)Wang, Cheng, Yu, Liu, Guo, Li, and
  Liu]{wang2025paperarena}
Daoyu Wang, Mingyue Cheng, Shuo Yu, Zirui Liu, Ze~Guo, Xin Li, and Qi~Liu.
\newblock Paperarena: An evaluation benchmark for tool-augmented agentic
  reasoning on scientific literature.
\newblock \emph{arXiv preprint arXiv:2510.10909}, 2025.
\newblock \doi{10.48550/arXiv.2510.10909}.
\newblock URL \url{https://arxiv.org/abs/2510.10909}.

\bibitem[Wang et~al.(2026)Wang, Jiang, Sun, Shi, Yu, Han, and
  Ji]{wang2026futurealigned}
Heng Wang, Pengcheng Jiang, Jiashuo Sun, Zhiyi Shi, Haofei Yu, Jiawei Han, and
  Heng Ji.
\newblock Learning to predict future-aligned research proposals with language
  models.
\newblock \emph{arXiv preprint arXiv:2603.27146}, 2026.
\newblock \doi{10.48550/arXiv.2603.27146}.
\newblock URL \url{https://arxiv.org/abs/2603.27146}.

\bibitem[Wu et~al.(2026{\natexlab{a}})Wu, Lu, Chen, Bragg, Yamada, Clark,
  Clifton, Torr, Zou, and Yu]{wu2026forecasting}
Sean Wu, Pan Lu, Yupeng Chen, Jonathan Bragg, Yutaro Yamada, Peter Clark,
  David~A. Clifton, Philip Torr, James Zou, and Junchi Yu.
\newblock Forecasting scientific progress with artificial intelligence.
\newblock \emph{arXiv preprint arXiv:2605.22681}, 2026{\natexlab{a}}.

\bibitem[Wu et~al.(2025)Wu, Pan, Xie, Zhou, Zhao, Ma, Du, Mao, Luu, and
  Wang]{wu2025antileakbench}
Xiaobao Wu, Liangming Pan, Yuxi Xie, Ruiwen Zhou, Shuai Zhao, Yubo Ma, Mingzhe
  Du, Rui Mao, Anh~Tuan Luu, and William~Yang Wang.
\newblock Antileakbench: Preventing data contamination by automatically
  constructing benchmarks with updated real-world knowledge.
\newblock In \emph{Proceedings of the 63rd Annual Meeting of the Association
  for Computational Linguistics (Volume 1: Long Papers)}, pages 18403--18419,
  2025.
\newblock \doi{10.18653/v1/2025.acl-long.901}.
\newblock URL \url{https://aclanthology.org/2025.acl-long.901/}.

\bibitem[Wu et~al.(2026{\natexlab{b}})Wu, Zhang, Li, Xu, Duan, Liu, Pan, Zhu,
  Zhou, Wei, Li, Chen, He, and Tan]{wu2026internatlas}
Yujun Wu, Dongxu Zhang, Xinchen Li, Jinhang Xu, Yiling Duan, Yumou Liu, Jiabao
  Pan, Qiyuan Zhu, Xuanhe Zhou, Jingxuan Wei, Siyuan Li, Jintao Chen, Conghui
  He, and Cheng Tan.
\newblock Intern-atlas: A methodological evolution graph as research
  infrastructure for {AI} scientists.
\newblock \emph{arXiv preprint arXiv:2604.28158}, 2026{\natexlab{b}}.
\newblock \doi{10.48550/arXiv.2604.28158}.
\newblock URL \url{https://arxiv.org/abs/2604.28158}.

\bibitem[Yamada et~al.(2025)Yamada, Lange, Lu, Hu, Lu, Foerster, Clune, and
  Ha]{yamada2025aiscientistv2}
Yutaro Yamada, Robert~Tjarko Lange, Cong Lu, Shengran Hu, Chris Lu, Jakob
  Foerster, Jeff Clune, and David Ha.
\newblock The {AI} scientist-v2: Workshop-level automated scientific discovery
  via agentic tree search.
\newblock \emph{arXiv preprint arXiv:2504.08066}, 2025.
\newblock \doi{10.48550/arXiv.2504.08066}.
\newblock URL \url{https://arxiv.org/abs/2504.08066}.

\bibitem[Yao et~al.(2023)Yao, Zhao, Yu, Du, Shafran, Narasimhan, and
  Cao]{yao2023react}
Shunyu Yao, Jeffrey Zhao, Dian Yu, Nan Du, Izhak Shafran, Karthik Narasimhan,
  and Yuan Cao.
\newblock React: Synergizing reasoning and acting in language models.
\newblock In \emph{International Conference on Learning Representations
  (ICLR)}, 2023.
\newblock URL \url{https://openreview.net/forum?id=WE_vluYUL-X}.

\bibitem[Ye et~al.(2026)Ye, Chen, Tu, Liu, Xiong, Schmidgall, and
  Bitterman]{ye2026proofoftime}
Bingyang Ye, Shan Chen, Jingxuan Tu, Chen Liu, Zidi Xiong, Samuel Schmidgall,
  and Danielle~S. Bitterman.
\newblock Proof of time: A benchmark for evaluating scientific idea judgments.
\newblock \emph{arXiv preprint arXiv:2601.07606}, 2026.

\bibitem[Ye et~al.(2024)Ye, Hu, Deng, Huang, Ma, Zhu, and Wang]{ye2024mirai}
Chenchen Ye, Ziniu Hu, Yihe Deng, Zijie Huang, Mingyu~Derek Ma, Yanqiao Zhu,
  and Wei Wang.
\newblock {MIRAI}: Evaluating {LLM} agents for event forecasting.
\newblock \emph{arXiv preprint arXiv:2407.01231}, 2024.
\newblock \doi{10.48550/arXiv.2407.01231}.
\newblock URL \url{https://arxiv.org/abs/2407.01231}.

\bibitem[Yuan et~al.(2025)Yuan, Ding, and Vlachos]{yuan2025forecast}
Zhangdie Yuan, Zifeng Ding, and Andreas Vlachos.
\newblock Introducing {FOReCAst}: The future outcome reasoning and confidence
  assessment benchmark.
\newblock In \emph{Advances in Neural Information Processing Systems},
  volume~38, 2025.
\newblock \doi{10.52202/085713-0808}.
\newblock URL
  \url{https://proceedings.neurips.cc/paper_files/paper/2025/hash/2277c7c3b112fcc6031a6f0d832df2a0-Abstract-Datasets_and_Benchmarks_Track.html}.
\newblock Datasets and Benchmarks Track.

\bibitem[{Z.AI}(2025)]{zai2025glm46}
{Z.AI}.
\newblock {GLM-4.6}.
\newblock \url{https://docs.z.ai/guides/llm/glm-4.6}, 2025.
\newblock Accessed 2026-09-18.

\bibitem[Zeng et~al.(2026)Zeng, Liu, Chen, He, Liao, Tian, Wang, Wang, Yang,
  Yin, Yin, Zhu, Cai, Chen, Chen, Du, Gao, Guo, Hu, Jiao, Li, Liu, Ni, Wen,
  Zhang, Zhang, Zhou, Blanchet, Qiu, Wang, and Huang]{zeng2025futurex}
Zhiyuan Zeng, Jiashuo Liu, Siyuan Chen, Tianci He, Yali Liao, Yixiao Tian,
  Jinpeng Wang, Zaiyuan Wang, Yang Yang, Lingyue Yin, Mingren Yin, Zhenwei Zhu,
  Tianle Cai, Zehui Chen, Jiecao Chen, Yantao Du, Xiang Gao, Jiacheng Guo,
  Liang Hu, Jianpeng Jiao, Xiangsheng Li, Jingkai Liu, Shuang Ni, Zhoufutu Wen,
  Ge~Zhang, Kaiyuan Zhang, Xin Zhou, Jose Blanchet, Xipeng Qiu, Mengdi Wang,
  and Wenhao Huang.
\newblock {FutureX}: An advanced live benchmark for {LLM} agents in future
  prediction.
\newblock In \emph{International Conference on Learning Representations}, 2026.
\newblock URL \url{https://openreview.net/forum?id=z28PLIEj6l}.

\bibitem[Zhao et~al.(2024)Zhao, Brumbaugh, Wang, Hajishirzi, and
  Smith]{zhao2024setclock}
Bowen Zhao, Zander Brumbaugh, Yizhong Wang, Hannaneh Hajishirzi, and Noah~A
  Smith.
\newblock Set the clock: Temporal alignment of pretrained language models.
\newblock In \emph{Findings of the Association for Computational Linguistics:
  ACL 2024}, pages 15015--15040, 2024.

\bibitem[Zhu et~al.(2026)Zhu, Zhang, Nie, and Zhang]{zhu2026sciimpact}
Hangxiao Zhu, Yuyu Zhang, Ping Nie, and Yu~Zhang.
\newblock {SciImpact}: A multi-dimensional, multi-field benchmark for
  scientific impact prediction.
\newblock In \emph{Findings of the Association for Computational Linguistics:
  ACL 2026}, pages 28923--28936. Association for Computational Linguistics,
  2026.
\newblock \doi{10.18653/v1/2026.findings-acl.1445}.
\newblock URL \url{https://aclanthology.org/2026.findings-acl.1445/}.

\end{thebibliography}
\bibliographystyle{plainnat}
\endgroup

\clearpage
\appendix
\etocdepthtag.toc{appendix}
\etocsettagdepth{main}{none}
\etocsettagdepth{appendix}{subsubsection}
\renewcommand{\contentsname}{Appendix Contents}
\tableofcontents
\clearpage

\setcounter{table}{0}
\setcounter{figure}{0}

\renewcommand{\thetable}{A\arabic{table}}
\renewcommand{\thefigure}{A\arabic{figure}}

\renewcommand{\theHtable}{appendix.table.\arabic{table}}
\renewcommand{\theHfigure}{appendix.figure.\arabic{figure}}

\setlength{\textfloatsep}{16pt plus 2pt minus 2pt}
\setlength{\intextsep}{12pt plus 2pt minus 2pt}
\makeatletter
\setlength{\@fpsep}{12pt plus 2pt minus 2pt}
\makeatother

\section{Limitations}
\label{sec:appendix-limitations}

\paragraph{Scope and transfer.}
ForeSci is intentionally a cutoff-controlled benchmark for forward-looking research judgement.
The current release covers four fast-moving AI areas, four decision families, and publicly
documented evidence. This design keeps each task historically auditable and fixes the information
boundary at which the decision is made. The reported results therefore characterize performance
under this evidence regime; extending the benchmark to additional domains, languages, forecast
horizons, or less documented settings provides a natural transfer test.

\paragraph{Research-question boundary.}
ForeSci focuses on fast-growing AI areas, where changes in methods, bottlenecks, evaluation
practices, and venue expectations provide observable temporal structure for a forward-looking
planning problem. It uses a graph-structured temporal taxonomy and its derived method-development
signals, bottleneck signals, feasibility and risk notes, venue-community metadata, and cutoff-local
paper evolution as proxy signals for prospective planning. The benchmark therefore asks whether an
agent can convert cutoff-visible evidence into decisions that align with subsequent developments
under this controlled setting.
This is a boundary of the research question, rather than a claim that slower-growing areas lack
scientific value, research importance, or a need for careful planning.

\FloatBarrier

\section{Responsible Research and Reproducibility}
\label{sec:appendix-responsible-research}

\paragraph{Artifacts, licenses, and intended use.}
ForeSci releases benchmark tasks, evaluation goldsets, prompts, schemas, scripts, and a
cutoff-aligned scholarly knowledge base for comparing research-assistant systems under temporal
controls. Source papers and bibliographic records retain their original access conditions;
derived artifacts are subject to the repository and release terms.

\begin{table}[H]
\centering
\scriptsize
\setlength{\tabcolsep}{3.5pt}
\caption{ForeSci release manifest. Evaluation-only targets are physically and logically separated
from model-visible prediction-time artifacts.}
\label{tab:artifact-manifest}
\begin{tabular}{p{0.28\linewidth}p{0.63\linewidth}}
\toprule
Artifact group & Released contents \\
\midrule
Prediction-time package & Public task files, cutoff and horizon metadata, and cutoff-aligned offline knowledge base \\
Evaluation package & Evaluation-only hidden targets and goldsets, metric prompts, schemas, and evaluator scripts \\
Construction package & Corpus filtering, taxonomy induction, task construction, and update scripts \\
Audit package & Run manifests, file checksums, aggregate validation summaries, and result-building scripts \\
\bottomrule
\end{tabular}
\end{table}

\paragraph{Privacy and content review.}
The benchmark uses public scholarly artifacts. Released tasks contain public question text and
generation metadata; hidden evaluation targets are packaged separately. Human-validation results
are reported in aggregate, with individual annotator records withheld.

\paragraph{Compute and model access.}
Answer generation uses inference through hosted or locally served API-compatible endpoints.
The learned evaluator consists of the family-specific XGBoost regressors described in
Section~\ref{sec:evaluation-metrics}. Methods reuse the offline knowledge base and retrieval
indexes, with fixed generation prompts, retrieval settings, and evaluation rubrics. Hidden future
targets are reserved for evaluation and excluded from method hyperparameter tuning.

\FloatBarrier

\section{Benchmark Construction and Temporal Integrity}
\label{sec:appendix-benchmark-artifacts}

The construction pipeline in Figure~\ref{fig:benchmark-construction} comprises corpus filtering,
temporal taxonomy induction, evidence extraction, and task curation. The following sections
describe these stages and the separation of prediction-time and evaluation artifacts.

\subsection{Corpus Filtering and Temporal Freezing}

For each domain, we start with broad domain queries, harvest papers through March 2026, normalize
metadata, and deduplicate query hits. The first pass is recall-oriented: it retains candidates with
non-trivial domain evidence even when the terminology is not yet stable.

The domain-relevance screen considers whether the paper's main problem, method, system, evaluation,
dataset, or application setting is substantively tied to the target area. The benchmark-core screen
then marks a paper as \textit{core} when it has a central domain contribution, a concrete research
asset type, and useful future-facing signal for tasks, methods, evaluation, bottlenecks, or design
patterns. We keep weaker relevant papers as \textit{support}; borderline cases may be retained for
audit; noisy or out-of-domain papers are excluded.

Public questions and accessible support are frozen at
historical cutoffs, while later papers are withheld for validation only. ForeSci uses
three-month and six-month forecast settings, and venue-conditioned tasks follow venue-cycle timing
because conference evidence appears on venue-specific schedules.

\subsection{Cutoff Slicing and Forecast-Oriented Taxonomy Adaptation}

Taxonomy induction uses the cutoff-aligned core/support corpora described in
Section~\ref{sec:data-collection}. To preserve temporal structure, papers are processed in
chronological slices. Earlier periods use coarser slices, while periods close to the cutoff use
finer slices when short-horizon movement matters. This design keeps recent changes visible instead
of smoothing them into the older literature.

Our taxonomy builder follows TaxoAdapt's multidimensional routing and adaptive expansion principle
\citep{kargupta2025taxoadapt}. Papers are routed across contribution dimensions such as tasks,
methods, datasets, evaluation methods, and application domains. Dense or poorly covered regions
trigger width or depth expansion, allowing new research subdirections to appear as the
cutoff-visible literature evolves.

We adapt this process for ForeSci in three ways: induction uses filtered core/support
papers, temporal slicing emphasizes cutoff-local deltas, and induced nodes must be grounded in
node evidence records before they can support benchmark construction.

\subsection{Candidate Direction Selection}

Candidate directions are selected from taxonomy nodes and small groups of related nodes. We retain a
candidate when it satisfies four criteria: it has sufficient pre-cutoff support, it expresses a
clear research decision, it is specific enough for evaluation, and it can be separated from hidden
future validation evidence. Candidates are revised or removed when they are ambiguous, duplicated,
too broad, too narrow, or weakly grounded in the underlying papers.

\subsection{Method-Development and Bottleneck Signals}

\emph{Method-development signals} are derived from method and evaluation
nodes, paper co-assignments, title/abstract method surfaces, full-text evidence, and
bottleneck--mechanism cues. They record relations such as extension, adaptation, replacement,
component reuse, and method competition. These signals provide trajectory evidence for comparing
directions and reasoning about mechanism-level change.

\emph{Bottleneck signals} summarize recurring limitations, evaluation gaps, reliability or safety
concerns, dataset or benchmark needs, and technical risks. We verify them against full-text evidence
from pre-cutoff papers, with emphasis on limitations that recur across multiple sources or connect
to concrete evaluation failures.

\subsection{Venue-Community Profiles}

Venue-conditioned tasks use \emph{venue-community signals from metadata}. We construct profiles for
venue families such as ACL/EMNLP/NAACL, ICLR/ICML/NeurIPS, AAAI/IJCAI, SIGIR/KDD, and
CVPR/ICCV/ECCV. Each profile summarizes contribution styles, maturity expectations, reviewer risks,
evidence-package expectations, and nearby compatible venue families. These profiles support
venue-cycle judgments based on research fit, evidence standards, and reviewer expectations.

\subsection{Human--LLM Collaborative Audit}

LLMs draft construction records from cutoff-aligned evidence by summarizing supporting papers,
extracting claims and limitations, identifying method-development and bottleneck signals, proposing
candidate directions, and flagging possible leakage risks.

Human experts then inspect the source papers and full-text evidence. They verify support strength,
temporal validity, specificity, and leakage risk; revise unclear wording; remove weak or duplicated
items; and approve the final construction records. The same audit process is applied before task
release.

\subsection{Task Curation and Artifact Separation}

Task curation checks three properties: a stable historical premise, a clear public decision, and
post-cutoff evidence suitable for validation. Items are revised when the decision is underspecified,
the validation evidence does not match the requested judgment, or multiple items express the same
research decision.

Each released task contains the public question $q$, cutoff $t$, forecast window, task-family
instructions, answer requirements, and pre-cutoff support packet $\mathcal{E}_{\leq t}(q)$.
Internal taxonomy identifiers, representative-paper lists, construction traces, and audit notes
remain internal. Hidden future validation targets $\mathcal{G}_{>t}(q)$ are held exclusively for
the evaluation protocol.
\begin{figure}[!htbp]
\centering
\includegraphics[width=0.8\linewidth]{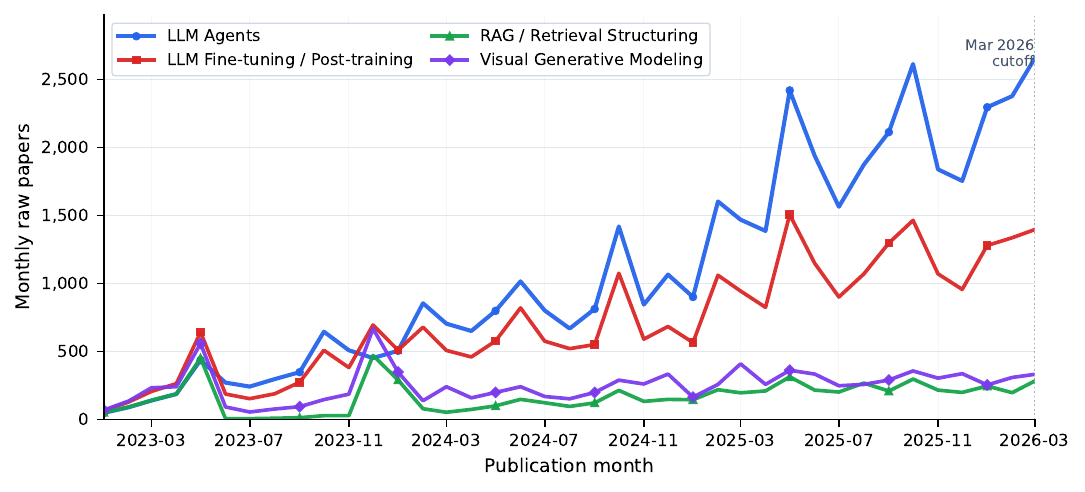}
\caption{Monthly raw-paper volume in the four ForeSci domains. Counts are unique papers
after normalizing paper identifiers within each domain-month. All four domains are shown from
January 2023 through the March 2026 corpus collection endpoint.}
\label{fig:monthly-raw-paper-counts}
\end{figure}

\begin{figure}[!htbp]
\centering
\includegraphics[width=0.8\linewidth]{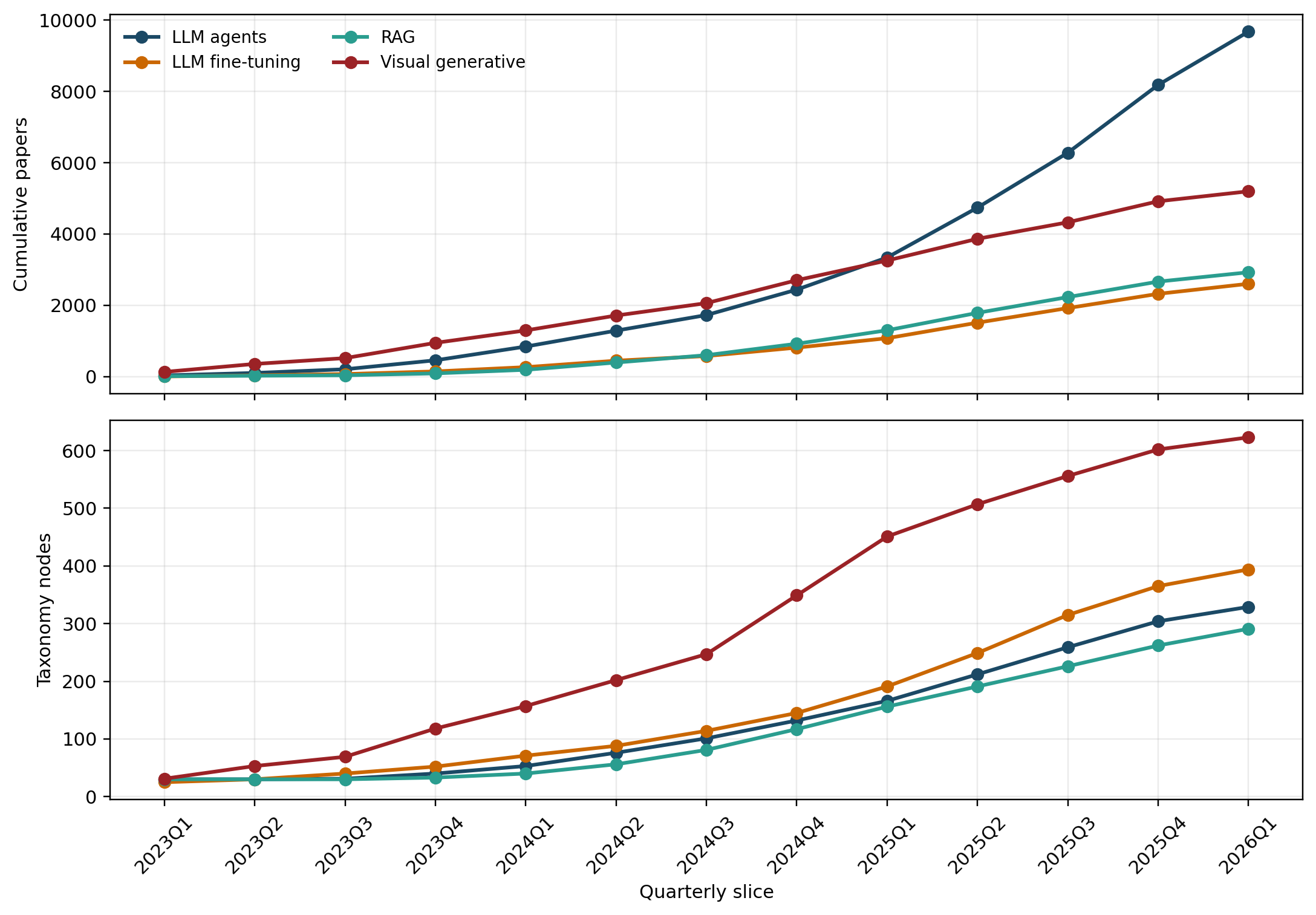}
\caption{Temporal evolution of the domain corpora and induced taxonomies used in
ForeSci. The curves summarize cumulative domain KB coverage and taxonomy-node growth over
the literature slices used for the release.}
\label{fig:taxonomy-evolution}
\end{figure}

\begin{figure*}[t]
\centering
\includegraphics[width=\textwidth]{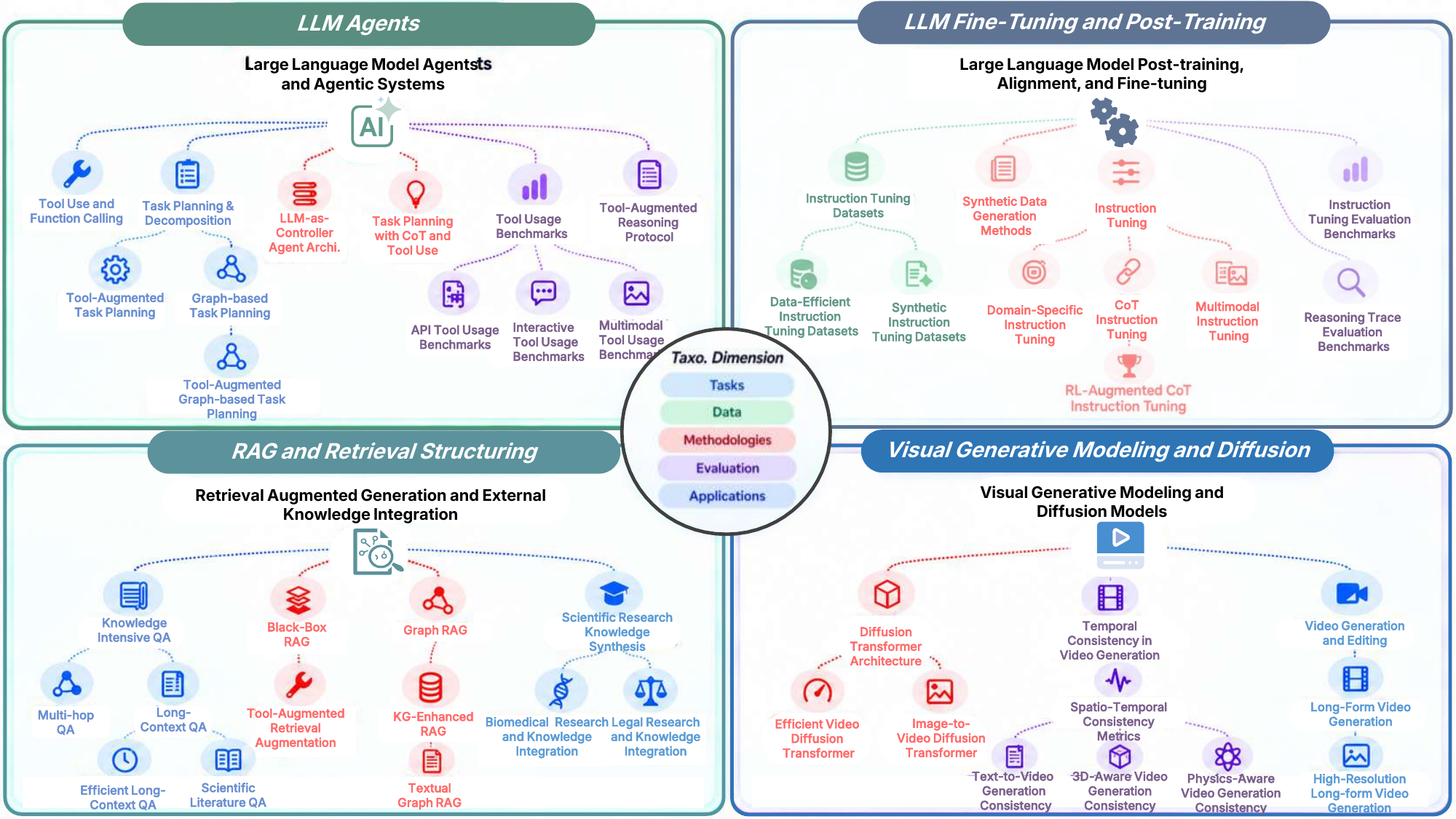}
\caption{Illustrative examples of temporal taxonomy branching in ForeSci. The figure
shows how broad pre-cutoff topics split into task-seeding subdirections; Table~\ref{tab:taxonomy-branches}
summarizes representative branch examples in text form.}
\label{fig:taxonomy-examples}
\end{figure*}

\begin{figure*}[t]
\centering
\includegraphics[width=\textwidth]{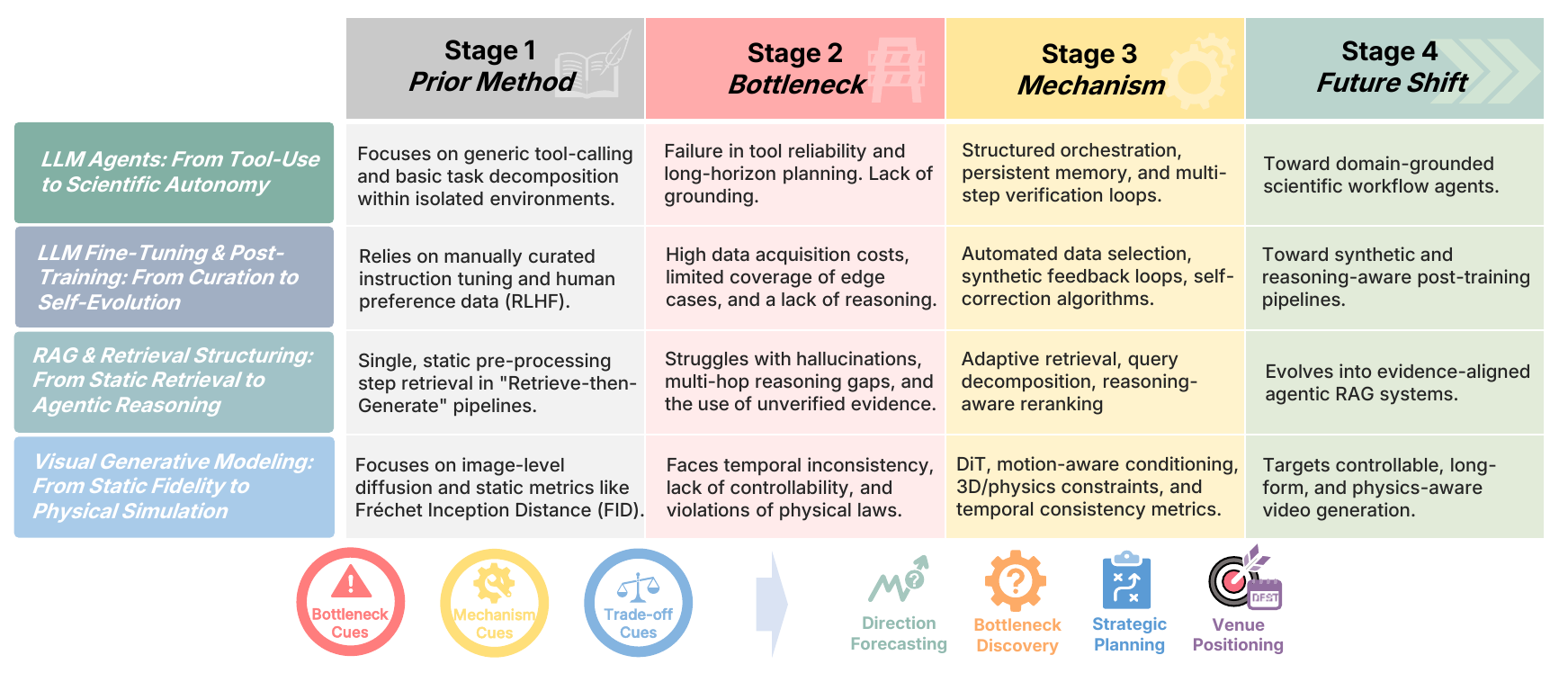}
\caption{Method-evolution signals for foresight task construction. ForeSci models each domain
as an evolutionary chain in which limitations
of prior methods expose bottleneck cues, emerging technical responses provide mechanism cues, and
the resulting bottleneck--mechanism interaction points toward a future research shift. The examples
show this progression for LLM agents, LLM fine-tuning and post-training, RAG and retrieval
structuring, and visual generative modeling. The bottom row summarizes how bottleneck, mechanism,
and trade-off cues feed the four downstream task families: direction forecasting,
bottleneck--opportunity discovery, strategic research planning, and venue-conditioned positioning.}
\label{fig:method-evolution}
\end{figure*}

\begin{table*}[t]
\centering
\scriptsize
\setlength{\tabcolsep}{3.0pt}
\caption{Representative taxonomy subtrees and method-evolution signals used during
ForeSci construction. The taxonomy column illustrates how temporally induced domain
structures refine broad research areas into fine-grained subdirections. The method-evolution column
summarizes the complementary bottleneck--mechanism--shift patterns used to seed future-facing task
targets.}
\label{tab:taxonomy-branches}
\begin{tabular}{p{0.18\linewidth}p{0.39\linewidth}p{0.36\linewidth}}
\toprule
Domain & Representative taxonomy subtree & Representative method-evolution signal \\
\midrule
LLM Agents &
Tool use and function calling $\rightarrow$ tool-augmented task planning $\rightarrow$ tool-augmented graph-based task planning;
tool usage benchmarks $\rightarrow$ API / interactive / multimodal tool usage benchmarks &
Generic tool use and task planning $\rightarrow$ tool reliability and long-horizon grounding bottlenecks $\rightarrow$ structured tool orchestration, memory, feedback, and verification loops $\rightarrow$ domain-grounded scientific workflow agents \\

LLM Fine-tuning and Post-training &
Instruction tuning datasets $\rightarrow$ data-efficient instruction tuning datasets / synthetic instruction tuning datasets;
instruction tuning $\rightarrow$ domain-specific, multimodal, and chain-of-thought instruction tuning &
Curated instruction and preference data $\rightarrow$ data cost, coverage, and weak reasoning-trace bottlenecks $\rightarrow$ data selection, synthetic feedback, self-correction, and process supervision $\rightarrow$ synthetic and reasoning-aware post-training pipelines \\

RAG and Retrieval Structuring &
Iterative retrieval-generation pipelines $\rightarrow$ retrieval strategy evaluation $\rightarrow$ adaptive / hybrid / multimodal retrieval evaluation;
reasoning-aware evaluation $\rightarrow$ evidence-aligned reasoning evaluation &
Retrieve-then-generate pipelines $\rightarrow$ hallucination, multi-hop grounding, stale evidence, and citation-fidelity bottlenecks $\rightarrow$ adaptive retrieval, query decomposition, evidence verification, and reasoning-aware reranking $\rightarrow$ evidence-aligned agentic RAG systems \\

Visual Generative Modeling and Diffusion &
Temporal consistency in video generation $\rightarrow$ spatio-temporal consistency metrics $\rightarrow$ text-to-video / 3D-aware / physics-aware consistency metrics;
video diffusion transformer architectures $\rightarrow$ efficient / image-to-video / video-editing diffusion transformers &
Image-level diffusion generation and static fidelity metrics $\rightarrow$ temporal inconsistency, controllability, physical plausibility, and long-form coherence bottlenecks $\rightarrow$ video diffusion transformers, motion-aware conditioning, 3D/physics constraints, and consistency metrics $\rightarrow$ controllable long-form and physics-aware video generation \\
\bottomrule
\end{tabular}
\end{table*}

Figure~\ref{fig:taxonomy-evolution} summarizes differences in corpus scale and taxonomy growth
across domains. The examples in Figure~\ref{fig:taxonomy-examples} and
Figure~\ref{fig:method-evolution} illustrate the subdirections and bottleneck--mechanism
relations used to construct tasks.

\subsection{Output-Level Temporal-Leakage Audit}
\label{sec:appendix-output-leakage-audit}

We audit all 5,000 outputs from the two proprietary backbones for explicit post-cutoff named
entities. Each output is bound to its own task cutoff, public question, and saved support snapshot.
A screening pass extracts paper, method, model, dataset, and system names and retains exact
surface-form overlaps with future targets. We then resolve each retained span to a canonical entity,
establish its earliest public date using the arXiv v1 date where available and otherwise an official
release or publisher first-online date, and attribute its possible origin to the public question, the
exact retrieved/full-text support snapshot, or parameter-origin risk. Support attribution requires
the entity to occur in the output's saved support snapshot. The audit concerns identifiable named
entities; conceptual similarity, shared mechanisms, and generic terminology fall outside its scope.

We label an entity as \emph{hard leakage} only when all five conditions hold:
\begin{enumerate}[leftmargin=*,nosep]
    \item the entity is an exact, locatable span in the answer;
    \item the span uniquely identifies a paper, method, model, dataset, or system;
    \item a reliable source dates its first public appearance after that answer's task cutoff;
    \item the entity is absent from the public question; and
    \item provenance adjudication, after checking the exact support snapshot, identifies either
    post-cutoff retrieval origin or unsupported parameter-origin risk rather than cutoff-safe prompt
    or support origin.
\end{enumerate}
We classify methods, modules, or directions explicitly proposed for development within the
forecast window as prospective proposals. Hard-leakage adjudication applies to uniquely
identifiable post-cutoff entities presented as already existing at the cutoff. The screen surfaces
57 mentions in 56 GPT-5.2 outputs and 60 mentions in 57 Gemini-3 outputs. Manual adjudication
finds no confirmed hard leakage under this definition: 0/2,500 for each backbone (95\% Wilson
upper bound 0.15\%) and 0/5,000 pooled (upper bound 0.08\%). These bounds describe detected
output-level leakage under the screening protocol; training-corpus contamination remains unmeasured.

\begin{table}[t]
\centering
\scriptsize
\setlength{\tabcolsep}{3.4pt}
\caption{Output-level temporal-leakage audit for proprietary backbones. Confirmed rates are
0/2,500 per backbone (95\% Wilson upper bound 0.15\%) and 0/5,000 pooled (upper bound 0.08\%).}
\label{tab:output-leakage-audit}
\begin{tabular}{lrrrr}
\toprule
Backbone & Outputs & Candidate mentions & Candidate outputs & Confirmed \\
\midrule
GPT-5.2 & 2,500 & 57 & 56 & 0 \\
Gemini-3 & 2,500 & 60 & 57 & 0 \\
\textbf{Pooled} & \textbf{5,000} & \textbf{117} & \textbf{113} & \textbf{0} \\
\bottomrule
\end{tabular}
\end{table}

\FloatBarrier

\section{Evaluation Protocol and Validation}
\label{sec:appendix-evaluation-details}

\subsection{Metric Calibration Details}
\label{sec:appendix-metric-details}

The following definitions specify the metrics introduced in Section~\ref{sec:evaluation}.

\paragraph{Prediction Factuality.}
Let $a$ denote a candidate answer and let $\mathcal{C}(a)=\{c_i\}_{i=1}^{m}$ be its extracted
atomic claims. The reported Prediction Factuality score is claim-level F1 over answer-claim support
and hidden claim-bank coverage. For each extracted answer claim $c_i$, a benchmark-aware verifier
assigns \textit{supported}, \textit{partially supported}, \textit{unsupported}, or
\textit{not checkable} relative to the public task and hidden claim units. Let
$\phi_i \in \{1,0.5,0,0\}$ be the corresponding answer-claim support score. Claim precision is
\begin{equation}
\mathrm{Prec}(a)=\frac{1}{m}\sum_{i=1}^{m}\phi_i.
\end{equation}
Let $\mathcal{G}^{+}=\{g_j\}_{j=1}^{n}$ denote the expanded hidden claim bank for the task. For each
hidden claim, the judge assigns \textit{covered}, \textit{partially covered}, or
\textit{not covered} by the candidate answer. Let $\psi_j \in \{1,0.5,0\}$ be the corresponding
coverage score. Claim recall is
\begin{equation}
\mathrm{Rec}(a)=\frac{1}{n}\sum_{j=1}^{n}\psi_j.
\end{equation}
The final score is
\begin{equation}
\mathrm{PredictionFactuality}(a)=
\frac{2\,\mathrm{Prec}(a)\,\mathrm{Rec}(a)}{\mathrm{Prec}(a)+\mathrm{Rec}(a)}.
\end{equation}
\paragraph{Future-Target Alignment (FTA).}
FTA is family-conditioned. For direction forecasting and bottleneck--opportunity tasks, we use
Reference-Guided FTA. The hidden future target is represented as a set of slots
$\mathcal{G}_{b}=\{g_j\}_{j=1}^{n}$. Each slot $g_j$ contains one or more acceptable textual
variants $V_j$, such as a primary future target, a root-bottleneck paraphrase, an unlocked
opportunity variant, or an evidence-backed mechanism variant. Variants are alternatives inside the
same target slot; they are not counted as additional targets. Negative confusions are retained for
qualitative auditing but are not counted as positive slots.

Let $\mathcal{C}(a)=\{c_i\}_{i=1}^{m}$ be the benchmark-relevant prediction claims extracted from
the answer. We embed every claim and target variant with bge-m3 and compute
$s(c,t)=\max(0,\cos(e(c),e(t)))$, placing pairwise similarity on a 0--1 scale. The slot score is the
best claim--variant match inside the slot:
\begin{equation}
S_j(a,b)=\max_{1\leq i\leq m,\;t\in V_j} s(c_i,t).
\end{equation}
The Reference-Guided FTA score is the mean slot score:
\begin{equation}
\mathrm{FTA}_{\mathrm{RG}}(a,b)=\frac{1}{n}\sum_{j=1}^{n} S_j(a,b).
\end{equation}
The target audit quantifies the breadth of the accepted variants. All 125 Direction
tasks contain acceptable neighbors (mean 4.43 per task). For Bottleneck, 94 of 125 tasks contain an
acceptable bottleneck variant (mean 3.44 when nonempty), and 86 contain an acceptable opportunity
variant (mean 4.06 when nonempty). All 125 Venue tasks contain acceptable venue-neighborhood entries
(mean 4.10). Across the 5,000 Direction and Bottleneck evaluations, 65.0\% use at least one
non-primary variant as a best slot match; 8.54\% of all slot-level best matches use such a variant.

For strategic planning, the hidden target is an ordered list
$\pi^*=(r_1,\ldots,r_K)$ over candidate research directions. The answer is parsed into an inferred
ranking $\hat{\pi}$. We score three deterministic components: whether the top item matches,
whether each preferred item appears in the same position or elsewhere, and whether pairwise order
relations are preserved. Let
\begin{align}
S_{\mathrm{top}} &= \mathbb{I}[\hat{\pi}_1=r_1],\\
S_{\mathrm{pos}} &= \frac{1}{K}\sum_{k=1}^{K} s_k,\\
S_{\mathrm{pair}} &= \frac{2}{K(K-1)}\sum_{i<j}\mathbb{I}[r_i \prec_{\hat{\pi}} r_j].
\end{align}
where $s_k=1$ when $r_k$ appears in position $k$, $s_k=0.5$ when it appears in another inferred
position, and $s_k=0$ when it is missing. The recall-style ranking score is the mean of these three
components. A symmetric precision-style score is computed over the inferred ranking using the same
top, position, and pairwise-order checks relative to $\pi^*$. The reported ranking-aware FTA is the
F1 of these precision and recall terms:
\begin{equation}
\mathrm{FTA}_{\mathrm{rank}}(a,b)=\frac{2PR}{P+R}.
\end{equation}
For Venue-Conditioned Positioning, we first canonicalize venue aliases within a fixed family
ontology (for example, individual members of a compatible venue family map to the same family).
Let $d$ and $\hat d$ denote the gold and inferred default families, let $A$ and $\hat A$ denote
the gold and inferred sets of conditional alternatives, and let $u$ and $\hat u$ denote the gold
and inferred contribution centers. We compute
\begin{equation}
S_{\mathrm{default}}=\mathbb{I}[\hat d=d],\qquad
S_{\mathrm{cond}}=\frac{|A\cap\hat A|}{|A|},\qquad
S_{\mathrm{contrib}}=\max_{v\in V_u}\max(0,\cos(e(v),e(\hat u))),
\end{equation}
where $V_u$ contains the accepted contribution-center variants and $e(\cdot)$ is the fixed
embedding function. The Venue score is
\begin{equation}
\mathrm{FTA}_{\mathrm{venue}}=0.50S_{\mathrm{default}}+0.20S_{\mathrm{cond}}+0.30S_{\mathrm{contrib}}.
\end{equation}
The three components measure default-family agreement, coverage of conditional alternatives, and
semantic alignment of the contribution center, respectively.
The reported FTA is $\mathrm{FTA}_{\mathrm{RG}}$ for Direction and Bottleneck tasks,
$\mathrm{FTA}_{\mathrm{rank}}$ for Planning, and the Venue-specific structured FTA for
Venue-Conditioned Positioning.

\paragraph{Evidence Traceability Score.}
For each task family $f$, we define exactly three decision slots $S_f$: forecast direction,
mechanism, and validation signal/risk for Direction; root bottleneck, causal bridge, and one-hop
opportunity for Bottleneck; top priority, ordering/dependency, and comparative ranking rationale for
Planning; and top venue, contribution profile, and cross-venue tradeoff for Venue. A first pass
extracts these slots from the final answer without judging correctness. For each extracted slot,
local bge-m3 retrieval selects the three most similar chunks from the output's attached pre-cutoff
support packet; the embedding similarity is used only for candidate recall and never enters the
score. A fixed language-model adjudicator applies the same structured rubric across families. It
labels support as direct, partial, topical-only, contradictory, or absent; labels the
answer's evidence-to-conclusion bridge as explicit, implicit, or absent; and identifies unresolved
conflict. For Venue, evidence need not literally mention the venue name: contribution profile,
evaluation regime, application setting, or technical emphasis is admissible only when the answer
explicitly connects it to the venue decision or cross-venue tradeoff.

For Venue we score three chain links: direct profile evidence with a bridge (profile evidence), direct or partial profile evidence
with an explicit connection to the selected venue (venue bridge), and direct evidence with an explicit
bridge for the cross-venue tradeoff (tradeoff evidence). Each link receives a continuous
traceability score in $[0,1]$.

Let $z_j\in[0,1]$ denote the continuous traceability score for decision slot $j$, reflecting
the strength of evidence support, the evidence-to-decision bridge, and unresolved conflict.
For Venue, $z_j$ scores the corresponding chain link defined above. The reported score is
the unweighted mean of the three continuous scores:
\begin{equation}
\mathrm{Trace}(a)=\frac{1}{3}\sum_{j=1}^{3} z_j.
\end{equation}
A stricter direct-support-plus-explicit-bridge criterion is retained as a separate Venue diagnostic.
Scoring uses the extracted decision slots and their pre-cutoff evidence, excluding hidden future
targets and system identity. Outputs without attached external support are reported as -- and
excluded from Trace aggregates. The released artifacts retain the underlying slot judgments.

\subsection{Human Annotation and PERS Modeling}
\label{sec:appendix-pers-annotation}

\paragraph{Reviewer Persuasiveness.}
Reviewer Persuasiveness estimates how convincing a research decision is to a reviewer. We
sample 1,200 answers from 120 tasks, balanced across the four families and stratified across
backbones and methods. Each answer is independently rated by three human annotators, who see the
public task and anonymized answer but not the hidden future target, support provenance, backbone, or
method. Each annotator assigns a holistic score from 1 (unusable) to 7 (exceptional). For answer
$a$, the human supervision target is the arithmetic mean of the three ratings,
\begin{equation}
y(a)=\frac{1}{3}\sum_{r=1}^{3}y_r(a).
\end{equation}

\begin{figure}[H]
\centering
\includegraphics[width=\textwidth]{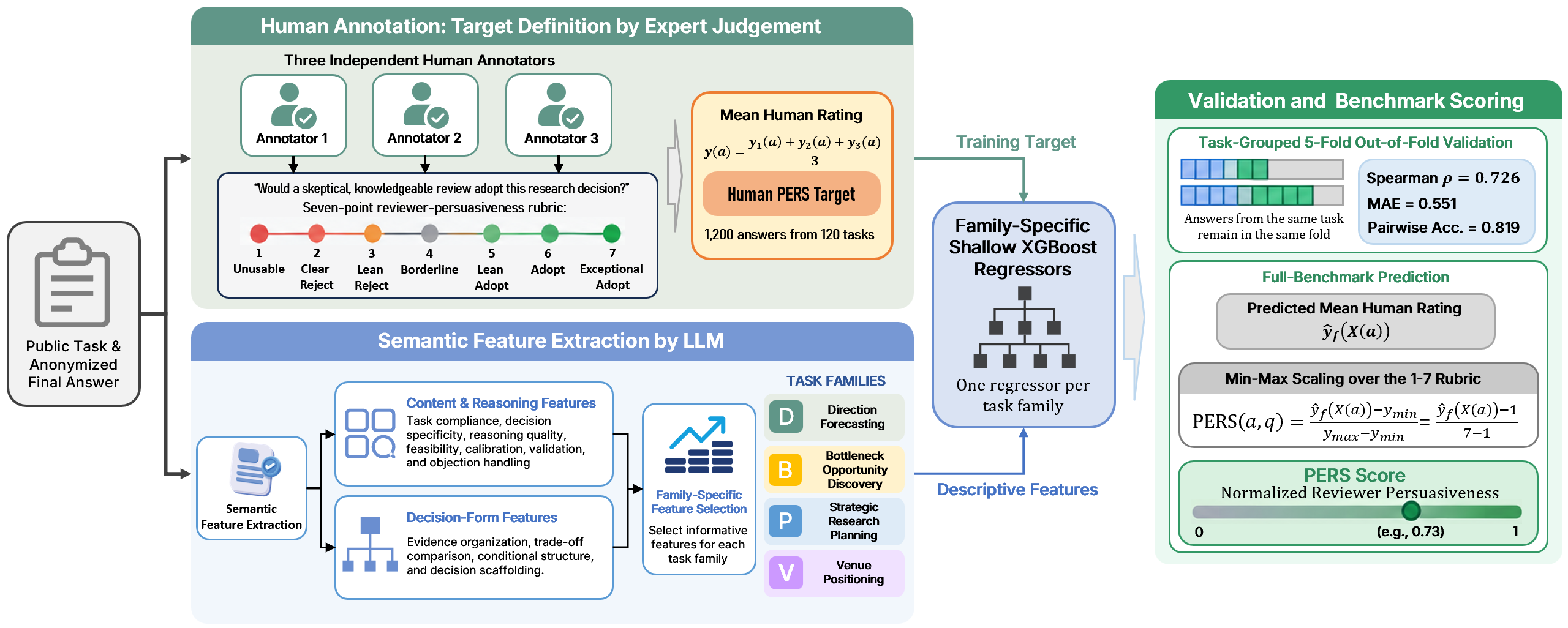}
\caption{Overview of PERS. Three independent human annotators score each anonymized answer on a
1--7 reviewer-persuasiveness rubric, and their mean rating defines the supervision target.
Semantic content, reasoning, and decision-form features extracted from the public task and answer
feed a shallow task-family-specific XGBoost regressor. Labeled answers receive task-grouped
out-of-fold predictions for validation; regressors fitted on the complete annotation set score the
remaining benchmark answers. Predicted ratings are min--max scaled using the rubric endpoints for
paper-facing PERS scores.}
\label{fig:pers-process}
\end{figure}

\paragraph{Human annotation protocol for $Y$.}
The human annotators received the following instructions. For each item, judge \emph{ex-ante
reviewer persuasiveness}: how willing you would be, as a skeptical and knowledgeable reviewer, to
adopt the answer's research decision based only on the public task and the reasoning visible in the
anonymized answer. Evaluate one complete answer at a time without comparing it with other answers.
Do not use outside information or attempt to infer the hidden future target, generation method, or
backbone. This judgment is distinct from future correctness, factuality, citation quality, evidence
traceability, and writing style. Length, citation count, polish, confidence, and technical
terminology receive no credit by themselves. Consider an unsupported factual statement only insofar
as it weakens the argument visible in the answer; do not externally verify it.

Make a holistic judgment of whether the answer states the required decision, provides concrete and
decision-relevant reasoning, distinguishes meaningful alternatives when required, addresses
feasibility and uncertainty, and gives a defensible basis for action. Before assigning a score,
identify the strongest adoption-relevant support and the single most decisive weakness. Your rating
is not an average of component subscores. Use the following seven anchors:
\begin{itemize}[leftmargin=*,nosep]
    \item 1 (unusable): no judgeable decision, severe instruction failure, or reasoning that cannot support adoption;
    \item 2 (clear reject): a decision is present, but decisive omissions, contradictions, or unsupported leaps dominate;
    \item 3 (lean reject): some plausible reasoning, but major unresolved weaknesses make adoption unlikely;
    \item 4 (borderline): an adequate but mixed case on which a reasonable reviewer could go either way;
    \item 5 (lean adopt): persuasive overall, with identifiable but non-decisive reservations;
    \item 6 (adopt): strong, concrete, well defended, and decision-ready, with only limited weaknesses;
    \item 7 (exceptional adopt): unusually compelling and robust, reserved for rare answers.
\end{itemize}
Mark an answer unjudgeable only when corruption or a fundamental instruction failure prevents
meaningful evaluation; weak but understandable answers remain judgeable.

All three PERS annotators either held or were pursuing doctoral degrees; the panel comprised
university associate professors and PhD candidates. Collectively, their research backgrounds span
LLM agents, video generation, and post-training.

We fit a separate shallow XGBoost regressor for each task family because the evidence needed for a
persuasive forecast, bottleneck diagnosis, research plan, and venue decision differs substantively.
The predictors use semantic features extracted from the public task and final answer, covering task
compliance, decision specificity, reasoning, feasibility, calibration, validation, evidence
presentation, and decision structure. Family-specific subsets are fixed before final evaluation.
The feature vector excludes backbone, generation method, answer length, family indicators, and
hidden targets; task family selects the regressor. Tables~\ref{tab:pers-content-features} and~\ref{tab:pers-form-features} give the
complete predictor sets and their operational meanings; each listed row corresponds to one model
input variable.

\begin{table*}[!htbp]
\centering
\scriptsize
\setlength{\tabcolsep}{2.4pt}
\renewcommand{\arraystretch}{1.06}
\caption{Content and reasoning variables used by the family-specific Reviewer Persuasiveness
regressors. Each row is one predictor; a checkmark indicates that the variable enters the
corresponding family model. B, D, P, and V denote Bottleneck--Opportunity Discovery, Direction
Forecasting, Strategic Research Planning, and Venue-Conditioned Positioning, respectively.}
\label{tab:pers-content-features}
\begin{tabular}{@{}p{0.245\textwidth}p{0.455\textwidth}ccccc@{}}
\toprule
Predictor & Operational meaning & Range & B & D & P & V \\
\midrule
Hard-constraint compliance & Required cardinality, cutoff, horizon, candidate-set, and exclusion constraints are obeyed. & 0--2 & $\checkmark$ & $\checkmark$ &  & $\checkmark$ \\
Required-output completeness & All explicitly requested decision outputs are present. & 0--2 &  &  &  & $\checkmark$ \\
Technical-object specificity & The decision identifies a concrete technical object rather than only a broad topic. & 0--2 & $\checkmark$ & $\checkmark$ &  & $\checkmark$ \\
Closest-alternative discrimination & The proposed root cause is distinguished from a symptom or competing explanation. & 0--2 & $\checkmark$ &  &  &  \\
Time-bounded plausibility & The forecast, opportunity, or plan is plausible within the stated horizon. & 0--2 & $\checkmark$ & $\checkmark$ & $\checkmark$ &  \\
Consequential capability unlocked & Resolving the bottleneck directly unlocks a consequential one-hop capability. & 0--2 & $\checkmark$ &  &  &  \\
Premises or signals stated & The answer exposes the signals on which its forecast rests. & 0--2 &  & $\checkmark$ &  &  \\
Premise-to-decision bridge & The answer connects its premises to the final decision. & 0--2 &  &  &  & $\checkmark$ \\
Shared-tradeoff discrimination & Alternatives are compared on common decision-relevant criteria. & 0--2 &  &  &  & $\checkmark$ \\
Decision-changing condition & A concrete condition is given under which the decision changes. & 0--2 &  &  & $\checkmark$ & $\checkmark$ \\
Dependencies and constraints & Material dependencies or implementation constraints are addressed. & 0--2 &  &  & $\checkmark$ &  \\
Observable validation signal & The answer specifies an observable outcome or measurement. & 0--2 &  &  & $\checkmark$ &  \\
Discriminative validation test & The proposed test can distinguish competing decisions or hypotheses. & 0--2 &  &  & $\checkmark$ &  \\
Inferential validity & Granting the stated premises, the conclusion follows without a decisive logical jump. & 0--5 & $\checkmark$ & $\checkmark$ &  &  \\
Causal-mechanism depth & The answer makes the intermediate how/why links in its causal account explicit. & 0--5 & $\checkmark$ & $\checkmark$ &  &  \\
Support relevance & Visible support bears directly on the claimed root cause and opportunity. & 0--5 & $\checkmark$ &  &  &  \\
Scope-boundary discipline & The claim is bounded to the settings for which its reasoning applies. & 0--5 & $\checkmark$ &  & $\checkmark$ &  \\
Epistemic calibration & Certainty and scope do not exceed what the visible reasoning licenses. & 0--5 &  & $\checkmark$ &  &  \\
Feasibility realism & Resources, dependencies, maturity, and timing are treated realistically. & 0--5 &  &  & $\checkmark$ & $\checkmark$ \\
Implementation actionability & The decision implies a concrete and executable work package. & 0--5 &  &  & $\checkmark$ &  \\
Validation decision value & The proposed evidence can change a go/no-go, priority, or venue decision. & 0--5 &  &  & $\checkmark$ & $\checkmark$ \\
Information prioritization & The decisive recommendation and reasons are recoverable without rewarding verbosity. & 0--5 &  &  & $\checkmark$ &  \\
Objection resilience & The strongest likely objection is identified and substantively addressed. & 0--5 &  &  & $\checkmark$ & $\checkmark$ \\
\bottomrule
\end{tabular}
\end{table*}

\begin{table*}[!htbp]
\centering
\scriptsize
\setlength{\tabcolsep}{2.4pt}
\renewcommand{\arraystretch}{1.08}
\caption{Decision-form variables used by the family-specific Reviewer Persuasiveness regressors.
These variables encode the presence of task-relevant organizational and argumentative patterns,
not their quality: 0 denotes absent, 1 localized, and 2 recurring. Bottleneck and Direction use no
decision-form variables in their final predictor sets.}
\label{tab:pers-form-features}
\begin{tabular}{@{}p{0.245\textwidth}p{0.455\textwidth}ccccc@{}}
\toprule
Predictor & Operational meaning & Range & B & D & P & V \\
\midrule
Claim--anchor pairing & Decision claims are placed together with their supporting anchors. & 0--2 &  &  & $\checkmark$ & $\checkmark$ \\
Cross-source synthesis & Multiple sources are combined within a single argumentative unit. & 0--2 &  &  & $\checkmark$ & $\checkmark$ \\
Conditional branching & The answer organizes recommendations around explicit conditions or cases. & 0--2 &  &  & $\checkmark$ & $\checkmark$ \\
Epistemic qualification & Claims are accompanied by visible uncertainty or scope qualifications. & 0--2 &  &  & $\checkmark$ &  \\
Tradeoff comparison & Alternatives are explicitly contrasted in terms of costs and benefits. & 0--2 &  &  & $\checkmark$ &  \\
Shared-criterion agenda comparison & Planning candidates are compared under a common criterion. & 0--2 &  &  & $\checkmark$ &  \\
Dependency--priority chain & Dependencies are used to justify the ordering of research priorities. & 0--2 &  &  & $\checkmark$ &  \\
Milestone decision rule & A milestone is tied to a subsequent continue, stop, or reprioritize decision. & 0--2 &  &  & $\checkmark$ &  \\
Sectioned decision scaffold & The venue decision is presented through explicit functional sections. & 0--2 &  &  &  & $\checkmark$ \\
Objection--response pattern & A likely objection is paired with a response or mitigation. & 0--2 &  &  &  & $\checkmark$ \\
Parallel venue review & Candidate venues are discussed in a parallel comparison structure. & 0--2 &  &  &  & $\checkmark$ \\
Shared-criteria venue comparison & Candidate venues are evaluated using the same stated criteria. & 0--2 &  &  &  & $\checkmark$ \\
Reviewer simulation & The answer explicitly anticipates how a reviewer would evaluate the submission. & 0--2 &  &  &  & $\checkmark$ \\
Evidence-upgrade mapping & Proposed evidence upgrades are mapped to specific claims or review concerns. & 0--2 &  &  &  & $\checkmark$ \\
Pivot-trigger condition & A concrete trigger is tied to changing the preferred venue. & 0--2 &  &  &  & $\checkmark$ \\
\bottomrule
\end{tabular}
\end{table*}

Labeled answers receive task-grouped out-of-fold predictions; the remaining answers are scored by
models fitted on the complete annotation set.

\subsection{End-to-End Worked Example}
\label{sec:appendix-worked-example}

Task RFDIR-0001 asks for one concrete direction for tool-using LLM agents, using evidence available
through 2025-09-30, for the 2025-10-01 to 2026-03-31 window, together with one trajectory label from
\textsc{accelerating}, \textsc{fragmenting}, \textsc{steady}, or \textsc{cooling}. A GPT-5.2 ARIS
output receives only the public task and its exposed pre-cutoff support packet. Representative
support records include ThinkGeo (tool correctness, formatting, reasoning, and final-answer
metrics), RAS-Eval (real-world security evaluation and simulation-to-real fidelity gaps), work on
deployment security and behavioral steering, and Tool-R1 (tool-learning momentum and RL cost).

The answer forecasts \emph{accelerating} work on trace-first reliability and regression suites for
tool-using agents: replayable execution checks for tool validity, schema adherence, runtime success,
and dependency chains, supplemented by structured assessment of plan quality. It argues that such
evaluation is more immediate than costlier RL or broader multi-agent redesign. At evaluation time,
the hidden target exposes six slots and nine variants covering standardized tool-agent evaluation,
trajectory-aware and long-horizon assessment, realistic settings, reliability and security beyond
final success, the accelerating label, and acceptable neighboring formulations.

\begin{table}[!htbp]
\centering
\scriptsize
\setlength{\tabcolsep}{3.2pt}
\caption{A prediction-to-evaluation example. The system sees the public task and
pre-cutoff support only; hidden post-cutoff target slots are opened only by the evaluator.}
\label{tab:worked-example}
\begin{tabular}{lp{0.70\linewidth}}
\toprule
Metric & Computation for RFDIR-0001 \\
\midrule
Fact = 1.000 & Five extracted decision-critical claims attain precision and hidden-claim recall of 1.000 over eight gold claim units. \\
FTA = 0.638 & For each of six target slots, the evaluator takes the best bge-m3 match over the answer claims and the slot's acceptable variants, then averages the six values. \\
Trace = 0.842 & The three decision slots receive continuous traceability scores in $[0,1]$; their unweighted mean is 0.842. \\
PERS = 0.743 & The three human scores are 6, 5, and 5, giving a normalized mean target of 0.722; 0.743 is the task-grouped OOF regression estimate. \\
\bottomrule
\end{tabular}
\end{table}

\subsection{Validation Analyses}
\label{sec:appendix-validation-analyses}

We report agreement among the PERS annotators, task-grouped predictive validation, aggregate
method contrasts, and human validation of claim extraction.

\subsubsection{Human Annotator Agreement}

Across the three annotator pairs, mean Spearman correlation is $0.818$ and mean
quadratic-weighted kappa (QWK) is $0.821$. Exact
agreement is $59.0\%$ on the seven-point scale, while $98.5\%$ of paired ratings differ by at most
one point.

\begin{table}[!htbp]
\centering
\scriptsize
\setlength{\tabcolsep}{3.4pt}
\caption{Pairwise agreement among the three human annotators who supplied the PERS training
targets. Exact agreement requires identical 1--7 ratings; ``Within 1 point'' allows an absolute
difference of at most one.}
\label{tab:pers-human-agreement}
\begin{tabular}{lrrrr}
\toprule
Annotator pair & Spearman $\rho$ & QWK & Exact & Within 1 point \\
\midrule
1 vs. 2 & 0.825 & 0.831 & 59.7\% & 99.3\% \\
1 vs. 3 & 0.811 & 0.813 & 59.8\% & 97.3\% \\
2 vs. 3 & 0.819 & 0.819 & 57.6\% & 98.8\% \\
\textbf{Mean} & \textbf{0.818} & \textbf{0.821} & \textbf{59.0\%} & \textbf{98.5\%} \\
\bottomrule
\end{tabular}
\end{table}

\subsubsection{Reviewer Persuasiveness Model Validation}
\label{sec:appendix-rp-model-validation}

The grouped-OOF evaluation covers 1,200 answers from 120 tasks. All answers sharing a task are
assigned to the same fold, so every prediction comes from a model trained without any answer from
that task. Aggregating the five held-out folds yields Spearman $\rho=0.726$, MAE $0.551$ points on
the 1--7 annotation scale, and within-task pairwise accuracy $0.819$ against the mean human rating.

\subsubsection{Aggregate Method Contrasts}

Table~\ref{tab:aggregate-method-contrasts} summarizes differences between the mean of CoI-style,
ResearchAgent-style, and ARIS-style and each baseline, averaging equally across the four
backbones.

\begin{table}[!htbp]
\centering
\scriptsize
\setlength{\tabcolsep}{3.5pt}
\caption{Mean differences between agent methods and baselines.
Agent averages include all three agent methods and four backbones. Native Trace is not applicable.}
\label{tab:aggregate-method-contrasts}
\begin{tabular}{llr}
\toprule
Metric & Contrast & $\Delta$ \\
\midrule
Fact & Agent avg. $-$ Hybrid & 0.0155 \\
FTA & Agent avg. $-$ Hybrid & 0.0163 \\
Trace & Agent avg. $-$ Hybrid & 0.0509 \\
PERS & Agent avg. $-$ Hybrid & 0.0181 \\
Fact & Agent avg. $-$ Native & 0.0203 \\
FTA & Agent avg. $-$ Native & 0.0221 \\
PERS & Agent avg. $-$ Native & 0.0501 \\
\bottomrule
\end{tabular}
\end{table}

In Table~\ref{tab:metrics-split-by-family}, the best--second gap is
at most 0.02 in 12/16 backbone--family cells for Fact, 12/16 for FTA, 8/16 for Trace,
and 14/16 for PERS.

\subsubsection{Claim Extraction Human Validation}
\label{sec:appendix-claim-extraction-human-validation}

The claim-extraction audit covers 240 answers, with 60 per family
(Table~\ref{tab:claim-extraction-human-quality}). Human annotators assess precision, recall,
atomicity, decision relevance, and noise relative to the source answer. Precision ranges from
0.78 to 0.86 and recall from 0.70 to 0.76. Planning and Venue have lower atomicity and higher
noise than Bottleneck and Direction. These measurements characterize the extraction stage;
the subsequent factuality judgments are a separate evaluation step.

\begin{table}[!htbp]
\centering
\scriptsize
\setlength{\tabcolsep}{3.6pt}
\caption{Human validation of automatic decision-critical claim extraction.}
\label{tab:claim-extraction-human-quality}
\begin{tabular}{lrrrrrrr}
\toprule
Family & $n$ & Prec. & Rec. & $F_1$ & Atomic & Relevant & Noise \\
\midrule
Bottleneck & 60 & 0.84 & 0.72 & 0.78 & 88.5\% & 85.0\% & 11.2\% \\
Direction & 60 & 0.86 & 0.76 & 0.81 & 91.0\% & 89.5\% & 8.0\% \\
Strategic Planning & 60 & 0.79 & 0.70 & 0.74 & 82.3\% & 81.0\% & 16.5\% \\
Venue & 60 & 0.78 & 0.73 & 0.75 & 80.1\% & 83.0\% & 18.2\% \\
\bottomrule
\end{tabular}
\end{table}

\FloatBarrier

\subsection{Evaluation Prompt Templates}
\label{sec:appendix-prompts}

The following condensed templates specify the claim-extraction, factuality, and traceability
rubrics and their output schemas.

\begin{tcolorbox}[
  breakable,
  colback=gray!4,
  colframe=gray!70,
  title=Atomic Claim Extraction Prompt,
  width=\columnwidth
]
\small
You are extracting benchmark-relevant atomic factual claims from a candidate answer.

Inputs:
\begin{itemize}[itemsep=1pt]
\item Public task definition: \verb|{PUBLIC_TASK_JSON}|
\item Candidate answer: \verb|{CANDIDATE_ANSWER}|
\end{itemize}

Your task:
\begin{itemize}[itemsep=1pt]
\item Decompose the answer into a small set of atomic factual claims.
\item Keep only claims that could be grounded in a research-paper benchmark.
\item Prefer claims about research directions, mechanisms, bottlenecks, venue fit, baselines, experiments, risks, trajectory, and evidence signals.
\item Ignore purely stylistic statements, generic advice, and duplicates.
\item Return at most \verb|{MAX_CLAIMS}| claims.
\end{itemize}

Output requirements:
\begin{itemize}[itemsep=1pt]
\item Output JSON only.
\item Do not include explanations, markdown, or extra text.
\end{itemize}

Output format:

{\ttfamily\small
\{\par
\quad "claims": ["...", "..."]\par
\}\par
}
\end{tcolorbox}
\begin{center}
\small \textit{Prompt template used for atomic claim extraction in Prediction Factuality.}
\end{center}

\begin{tcolorbox}[
  breakable,
  colback=gray!4,
  colframe=gray!70,
  title=Claim-Level Factuality and Coverage Prompt,
  width=\columnwidth
]
\small
You are a claim-level evaluator for a research-foresight benchmark. Compute claim precision and
gold-claim recall for the candidate answer. This is a factual/content coverage metric, not a
research-taste metric.

Inputs:
\begin{itemize}[itemsep=1pt]
\item Public task definition: \verb|{PUBLIC_TASK_JSON}|
\item Candidate answer claims: \verb|{ANSWER_CLAIMS_JSON}|
\item Gold claim units: \verb|{GOLD_CLAIM_UNITS_JSON}|
\end{itemize}

Judgment rules:
\begin{itemize}[itemsep=1pt]
\item For answer claims, label each as \verb|supported|, \verb|partially_supported|, \verb|unsupported|, or \verb|not_checkable| relative to the public task and gold claim units.
\item For gold claim units, label coverage as \verb|covered|, \verb|partially_covered|, or \verb|not_covered| by the candidate answer.
\item Do not reward generic topical overlap when the mechanism, venue rationale, or decision is different.
\item Keep this metric independent of future-target alignment scoring.
\end{itemize}

Output format:

{\ttfamily\small
\{\par
\quad "answer\_claim\_verdicts": [\par
\quad\quad \{"claim": "...", "label": "supported | partially\_supported | unsupported | not\_checkable",\par
\quad\quad "matched\_gold\_claim\_ids": ["..."], "rationale": "..."\}\par
\quad ],\par
\quad "gold\_claim\_coverage": [\par
\quad\quad \{"gold\_claim\_id": "...", "label": "covered | partially\_covered | not\_covered",\par
\quad\quad "matched\_answer\_claims": ["..."], "rationale": "..."\}\par
\quad ]\par
\}\par
}
\end{tcolorbox}
\begin{center}
\small \textit{Prompt template used for claim-level factuality and coverage in Prediction Factuality.}
\end{center}

\begin{tcolorbox}[
  breakable,
  colback=gray!4,
  colframe=gray!70,
  title=Evidence Traceability Slot Extraction and Adjudication,
  width=\columnwidth
]
\small
Stage 1: extract the three family-specific decision slots from the candidate answer. Do not solve
the task or judge correctness. If a slot is absent, return an empty claim and mark it not present.

Stage 2: for each slot, judge how strongly the claim is supported by its supplied pre-cutoff
candidate evidence chunks.
\begin{itemize}[itemsep=1pt]
\item Use only the evidence content supplied for that slot; do not use outside knowledge or hidden future targets.
\item Label support as \verb|direct|, \verb|partial|, \verb|topical_only|, \verb|contradictory|, or \verb|none|.
\item Label the evidence-to-conclusion bridge as \verb|explicit|, \verb|implicit|, or \verb|absent|.
\item Mark unresolved conflict only when materially conflicting evidence is present and the answer does not resolve it.
\item Assign a continuous traceability score in $[0,1]$ based on evidence support, the evidence-to-decision bridge, and unresolved conflict.
\item Evidence quantity and embedding similarity do not add points.
\end{itemize}

Inputs:
\begin{itemize}[itemsep=1pt]
\item Public task family and question: \verb|{PUBLIC_TASK}|
\item Candidate answer: \verb|{CANDIDATE_ANSWER}|
\item Family-specific slot schema: \verb|{SLOT_SCHEMA}|
\item Top-3 candidate evidence chunks per extracted slot: \verb|{CANDIDATE_EVIDENCE}|
\end{itemize}

Output format:

{\ttfamily\small
\{\par
\quad "slot\_results": [\par
\quad\quad \{"slot\_id": "...", "support\_status": "direct | partial | topical\_only | contradictory | none",\par
\quad\quad "bridge\_status": "explicit | implicit | absent", "unresolved\_conflict": false,\par
\quad\quad "slot\_score": 0.0,\par
\quad\quad "evidence\_anchors": ["..."], "short\_reason": "..."\}\par
\quad ]\par
\}\par
}
\end{tcolorbox}
\begin{center}
\small \textit{Condensed two-stage prompt used for TRACE slot extraction and evidence-chain adjudication; Venue uses the profile-to-venue chain rule.}
\end{center}

\FloatBarrier

\section{System Adaptation and Comparison Controls}
\label{sec:appendix-agent-adaptations}

The three agent systems use offline adaptations that retrieve from the pre-cutoff knowledge base
and expose evidence artifacts for evaluation.

\paragraph{Shared adaptation principles.}
Across all agent baselines, we applied the same three benchmark-side constraints.
\begin{itemize}[noitemsep]
\item \textbf{Open-web removal.} All online literature search or browsing steps were replaced by
offline access to the benchmark knowledge base, including paper metadata, section-level snippets,
and benchmark-side support packets.
\item \textbf{Cutoff-faithful evidence flow.} Retrieval, intermediate reasoning, and final answer
rendering were constrained to use only pre-cutoff assets. Hidden future evidence remained strictly
evaluation-only.
\item \textbf{Benchmark-native output exposure.} Each adapted agent was modified to emit the
support artifacts needed by the benchmark judges, such as retrieval traces, evidence items, and
family-native structured answer fields when available.
\end{itemize}

\paragraph{Agent-specific adaptations.}
CoI-style retains multi-chain idea construction and replaces online literature search with an
offline KB retrieval adaptor. ResearchAgent-style gathers evidence through benchmark support
packets and uses task-specific planning and review prompts. ARIS-style uses offline hybrid
retrieval, task-specific evidence construction, and a final renderer that enforces the candidate
set and required answer fields.
All three agents render their final decisions using the task-family output schemas.

\FloatBarrier

\section{Supplementary Results and Evaluator Analyses}
\label{sec:appendix-extra-results}

\subsection{Evaluation Results by Task Family}
Table~\ref{tab:metrics-split-by-family} reports results by task family. Method rankings vary
across families, backbones, and metrics.

\begin{table*}[!htbp]
\centering
\tiny
\setlength{\tabcolsep}{2.0pt}
\renewcommand{\arraystretch}{0.90}
\caption{Main family-level results across four answer-generation backbones. Fact and FTA use the final evaluation outputs, Pers is the predicted mean three-annotator rating normalized to $[0,1]$, and Trace is the mean continuous evidence-traceability score across three task-specific decision slots, including the profile-to-venue links for Venue. Each family block reports five methods; bold marks the best method within the same family, backbone, and metric.}
\label{tab:metrics-split-by-family}
\resizebox{\textwidth}{!}{%
\begin{tabular}{@{}llrrrrrrrrrrrrrrrr@{}}
\toprule
& & \multicolumn{4}{c}{Qwen3-235B} & \multicolumn{4}{c}{GPT-5.2} & \multicolumn{4}{c}{GLM-4.6} & \multicolumn{4}{c}{Gemini-3} \\
\cmidrule(lr){3-6}\cmidrule(lr){7-10}\cmidrule(lr){11-14}\cmidrule(l){15-18}
Family & Method & Fact. & FTA & Trace & Pers. & Fact. & FTA & Trace & Pers. & Fact. & FTA & Trace & Pers. & Fact. & FTA & Trace & Pers. \\
\textbf{Bottleneck--Opportunity} & Native LLM & 0.840 & 0.608 & -- & 0.675 & 0.799 & \textbf{0.614} & -- & 0.735 & 0.627 & 0.574 & -- & \textbf{0.528} & 0.672 & 0.576 & -- & \textbf{0.691} \\
 & Hybrid RAG & 0.824 & \textbf{0.609} & \textbf{0.621} & 0.668 & 0.804 & 0.613 & 0.642 & 0.747 & 0.611 & \textbf{0.589} & 0.536 & 0.502 & 0.704 & 0.589 & \textbf{0.526} & 0.667 \\
 & CoI & 0.851 & 0.607 & 0.532 & 0.659 & 0.819 & 0.610 & 0.667 & \textbf{0.762} & 0.672 & 0.584 & 0.533 & 0.518 & 0.700 & 0.583 & 0.455 & 0.656 \\
 & ResearchAgent & 0.802 & 0.608 & 0.554 & 0.664 & \textbf{0.829} & 0.610 & \textbf{0.684} & 0.752 & 0.653 & 0.586 & \textbf{0.598} & 0.493 & 0.716 & \textbf{0.591} & 0.507 & 0.654 \\
 & ARIS & \textbf{0.860} & 0.607 & 0.521 & \textbf{0.694} & 0.813 & 0.607 & 0.584 & 0.751 & \textbf{0.702} & 0.588 & 0.511 & 0.509 & \textbf{0.735} & 0.576 & 0.417 & 0.644 \\
\midrule
\textbf{Direction Forecasting} & Native LLM & \textbf{0.598} & 0.635 & -- & 0.626 & \textbf{0.670} & 0.635 & -- & 0.710 & 0.498 & 0.581 & -- & 0.532 & 0.573 & 0.601 & -- & \textbf{0.609} \\
 & Hybrid RAG & 0.567 & \textbf{0.637} & \textbf{0.763} & 0.604 & 0.645 & 0.634 & 0.621 & 0.695 & 0.523 & \textbf{0.597} & 0.669 & 0.550 & 0.589 & \textbf{0.606} & 0.620 & 0.591 \\
 & CoI & 0.577 & 0.635 & 0.738 & 0.641 & 0.658 & \textbf{0.637} & \textbf{0.671} & 0.721 & \textbf{0.534} & 0.592 & 0.574 & \textbf{0.561} & \textbf{0.601} & 0.602 & 0.606 & 0.587 \\
 & ResearchAgent & 0.568 & 0.633 & 0.703 & 0.633 & 0.662 & 0.633 & 0.658 & 0.720 & 0.501 & 0.584 & 0.635 & 0.542 & 0.595 & 0.604 & 0.569 & 0.582 \\
 & ARIS & 0.542 & 0.627 & 0.705 & \textbf{0.649} & 0.589 & 0.628 & 0.595 & \textbf{0.723} & 0.512 & 0.591 & \textbf{0.684} & 0.557 & 0.573 & 0.600 & \textbf{0.669} & 0.585 \\
\midrule
\textbf{Strategic Planning} & Native LLM & 0.465 & 0.580 & -- & 0.480 & 0.441 & 0.548 & -- & 0.586 & 0.334 & 0.449 & -- & 0.372 & 0.421 & 0.554 & -- & 0.458 \\
 & Hybrid RAG & 0.446 & 0.554 & 0.495 & 0.602 & 0.446 & 0.553 & 0.473 & 0.734 & 0.411 & 0.521 & 0.536 & 0.525 & 0.422 & 0.521 & 0.459 & 0.641 \\
 & CoI & 0.504 & 0.644 & 0.479 & 0.655 & 0.476 & 0.582 & 0.528 & \textbf{0.767} & 0.461 & \textbf{0.646} & \textbf{0.542} & 0.558 & 0.414 & 0.528 & \textbf{0.512} & 0.663 \\
 & ResearchAgent & \textbf{0.532} & \textbf{0.687} & 0.486 & 0.647 & 0.481 & 0.577 & \textbf{0.582} & 0.758 & \textbf{0.482} & 0.629 & 0.538 & 0.542 & \textbf{0.424} & 0.544 & 0.480 & \textbf{0.668} \\
 & ARIS & 0.460 & 0.598 & \textbf{0.507} & \textbf{0.661} & \textbf{0.488} & \textbf{0.587} & 0.567 & \textbf{0.767} & 0.441 & 0.565 & 0.532 & \textbf{0.579} & 0.422 & \textbf{0.555} & 0.500 & 0.660 \\
\midrule
\textbf{Venue Positioning} & Native LLM & \textbf{0.534} & 0.712 & -- & 0.582 & 0.551 & 0.680 & -- & 0.720 & 0.542 & 0.738 & -- & 0.581 & 0.570 & 0.776 & -- & 0.648 \\
 & Hybrid RAG & 0.526 & 0.700 & 0.438 & 0.542 & 0.576 & 0.694 & 0.503 & 0.724 & 0.553 & 0.762 & 0.347 & 0.601 & 0.555 & 0.776 & 0.393 & 0.649 \\
 & CoI & 0.518 & 0.664 & 0.572 & \textbf{0.607} & 0.575 & 0.729 & 0.726 & \textbf{0.741} & 0.554 & \textbf{0.804} & 0.502 & 0.638 & 0.584 & 0.781 & 0.610 & 0.664 \\
 & ResearchAgent & 0.528 & 0.690 & 0.531 & 0.602 & 0.580 & 0.723 & 0.688 & 0.733 & \textbf{0.571} & 0.755 & 0.455 & \textbf{0.642} & \textbf{0.590} & \textbf{0.810} & 0.574 & \textbf{0.681} \\
 & ARIS & 0.523 & \textbf{0.731} & \textbf{0.797} & 0.591 & \textbf{0.596} & \textbf{0.771} & \textbf{0.823} & 0.729 & 0.511 & 0.749 & \textbf{0.708} & 0.604 & \textbf{0.590} & 0.772 & \textbf{0.750} & 0.588 \\
\bottomrule
\end{tabular}
}
\end{table*}

\subsection{Auxiliary Evaluator Stability}
\label{sec:appendix-scalar-metric-stability}

We report run-to-run stability on a Qwen3-235B subset of 100 tasks and five methods, giving
500 answers. Candidate answers are held fixed across five independent evaluation passes.
For each metric, Table~\ref{tab:scalar-metric-stability} reports the largest standard deviation
and largest range of method-level run means across all family--method cells. This auxiliary test
predates the current slot-based Trace and regression-based PERS formulations, so it is used
only to characterize the retained content metrics.

\begin{table}[t]
\centering
\scriptsize
\setlength{\tabcolsep}{3.2pt}
\caption{Auxiliary five-run stability of Fact and FTA on the Qwen3-235B 100-task subset. ``Max mean SD''
and ``Max mean range'' summarize method-level run means across family--method cells. ``Mean row
SD'' summarize per-row variation across the same five runs.}
\label{tab:scalar-metric-stability}
\begin{tabular}{lrrr}
\toprule
Metric & Max SD & Max range & Row SD \\
\midrule
Prediction Factuality & 0.021 & 0.053 & 0.045 \\
Future-Target Alignment & 0.006 & 0.014 & 0.004 \\
\bottomrule
\end{tabular}
\end{table}

FTA varies less across these repeated evaluations than Fact. The maximum range of method-level
means is 0.014 for FTA and 0.053 for Fact. PERS predictive validation is reported in
Appendix~\ref{sec:appendix-rp-model-validation}.

\subsection{Answer Lengths}
\label{sec:appendix-answer-lengths}

\begin{table*}[!htbp]
\centering
\scriptsize
\setlength{\tabcolsep}{3.8pt}
\caption{Answer-length profiles on aligned outputs. Lengths are regex-based word
counts. Each row contains 625 answers from one generator backbone and one task
family in the current aligned answer snapshot.}
\label{tab:answer-length-profiles}
\begin{tabular}{llrrrr}
\toprule
Generator & Family & Mean & Median & P90 & Max \\
\midrule
GLM-4.6 & Bottleneck--Opportunity & 124.9 & 121 & 161.0 & 228 \\
GLM-4.6 & Direction Forecasting & 97.7 & 96 & 120.0 & 211 \\
GLM-4.6 & Strategic Planning & 154.8 & 154 & 189.6 & 257 \\
GLM-4.6 & Venue Positioning & 236.6 & 232 & 296.0 & 380 \\
\midrule
GPT-5.2 & Bottleneck--Opportunity & 339.8 & 328 & 432.6 & 692 \\
GPT-5.2 & Direction Forecasting & 236.3 & 232 & 305.0 & 400 \\
GPT-5.2 & Strategic Planning & 258.9 & 256 & 293.6 & 364 \\
GPT-5.2 & Venue Positioning & 424.0 & 411 & 533.0 & 630 \\
\midrule
Gemini-3 & Bottleneck--Opportunity & 197.9 & 194 & 259.0 & 322 \\
Gemini-3 & Direction Forecasting & 155.5 & 155 & 182.0 & 215 \\
Gemini-3 & Strategic Planning & 204.1 & 205 & 227.0 & 263 \\
Gemini-3 & Venue Positioning & 316.0 & 285 & 424.2 & 524 \\
\midrule
Qwen3-235B & Bottleneck--Opportunity & 198.3 & 191 & 262.0 & 384 \\
Qwen3-235B & Direction Forecasting & 173.0 & 173 & 218.6 & 286 \\
Qwen3-235B & Strategic Planning & 172.9 & 174 & 196.0 & 227 \\
Qwen3-235B & Venue Positioning & 243.3 & 236 & 337.6 & 425 \\
\bottomrule
\end{tabular}
\end{table*}

Table~\ref{tab:answer-length-profiles} reports answer lengths for the formal evaluation outputs.
All four backbones share the same 2,500 \((\mathrm{task\_id}, \mathrm{method})\) keys, with
625 answers per backbone and family. In each family, GPT-5.2 has the largest mean word count
and GLM-4.6 the smallest. These statistics describe the edited outputs used for formal evaluation.

\subsection{Content-Metric Correlation}
\label{sec:appendix-metric-correlation}

We compute pairwise Pearson and Spearman correlations among Fact, FTA, and PERS over the
10,000 final evaluations. This analysis includes Native LLM outputs, for which Trace is
unavailable, and therefore covers the three metrics shared by all methods.

Overall Spearman correlation is 0.448 for Fact vs. FTA, 0.189 for Fact vs. PERS, and 0.089 for
FTA vs. PERS. Correlations vary by task family.
Planning and Venue show much stronger Fact-vs.-FTA coupling: Strategic Planning
has Spearman 0.816 and Venue Positioning has 0.710, compared with 0.292 for
Bottleneck--Opportunity and 0.373 for Direction Forecasting. In these tasks, factual claim correctness
and target alignment both depend on the priority or conditional positioning decision.
Venue FTA vs. PERS correlation is 0.098, indicating a weaker association between target
alignment and predicted reviewer persuasiveness.

\begin{figure*}[!htbp]
\centering
\includegraphics[width=0.92\textwidth]{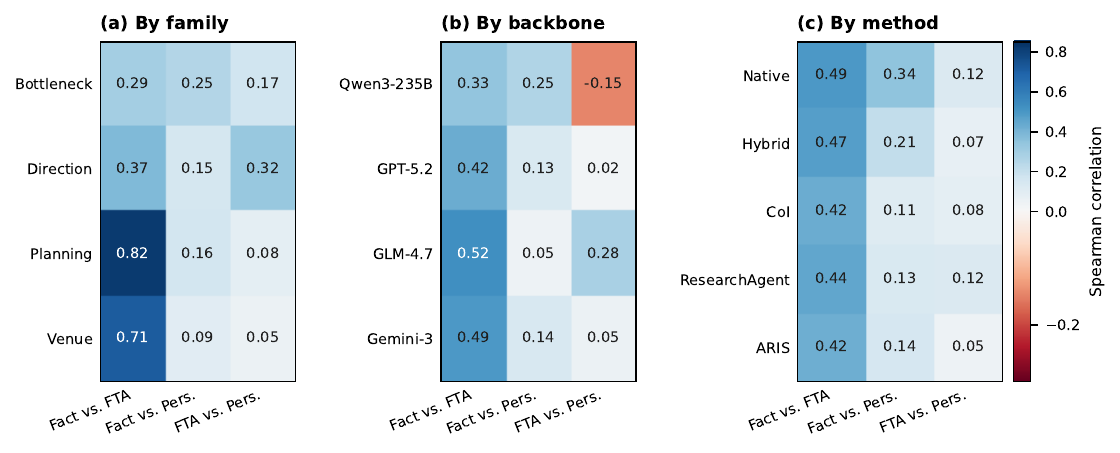}
\caption{Supplementary content-metric correlation diagnostic over 10,000 final benchmark
rows. Values are Spearman rank correlations among Prediction Factuality, Future-Target Alignment,
and regression-based Reviewer Persuasiveness. Trace is unavailable for Native LLM outputs and
is excluded from this all-method analysis. Correlations vary by family:
planning and venue tasks have much stronger Fact-vs.-FTA coupling than bottleneck and
direction tasks, while backbone and method differences are more moderate.}
\label{fig:metric-correlation-summary}
\end{figure*}

Across backbones,
Fact vs. FTA Spearman ranges from 0.368 for Qwen3-235B to 0.577 for GLM-4.6. FTA vs. PERS ranges
from $-0.180$ for Qwen3-235B to $-0.021$ for GLM-4.6. Across methods, Fact vs. FTA ranges
from 0.440 for CoI-style to 0.527 for Native LLM, while FTA vs. PERS ranges from $-0.052$
to 0.002. The pooled and subgroup correlations describe different aggregation levels and
should be read together.

\section{Error and Mechanism Diagnostics}
\label{sec:appendix-diagnostics}
\subsection{Detailed Error Analysis}
\label{sec:appendix-detailed-error-analysis}

The error analysis uses the 10,000 formal evaluations, their evaluator rationales, gold-claim
coverage summaries, future-target judgments, and support metadata. Drift annotations are joined
where available. For each metric, we calculate the global 20th percentile score threshold and
mark rows scoring strictly below this threshold as low scoring. Because of score ties at the
cutoff point, the actual percentage of selected low-scoring rows is less than 20\% (e.g.,
approximately 10.4\% for Pers and 16.9\% for Fact). Trace percentiles are calculated only over
evidence-grounded methods, as Native LLMs do not expose trace artifacts. These labels select
cases for inspection.

\begin{table*}[t]
\centering
\scriptsize
\setlength{\tabcolsep}{4.0pt}
\caption{Low-score rates used for the diagnostic error analysis. Rows scoring strictly below each
metric's global 20th percentile threshold are selected; ties at the threshold are excluded.
Trace thresholds and rates use evidence-grounded methods only; Native LLM traceability is
not applicable.}
\label{tab:error-low-counts}
\begin{tabular}{llrrrr}
\toprule
Grouping & Unit & Low Fact & Low FTA & Low Pers. & Low Trace \\
\midrule
\multirow{4}{*}{Family}
& Bottleneck--Opportunity & 0.062 & 0.013 & 0.174 & 0.308 \\
& Direction Forecasting & 0.184 & 0.004 & 0.012 & 0.164 \\
& Strategic Planning & 0.315 & 0.513 & 0.192 & 0.269 \\
& Venue Positioning & 0.114 & 0.251 & 0.037 & 0.018 \\
\midrule
\multirow{5}{*}{Method}
& Native LLM & 0.186 & 0.215 & 0.187 & -- \\
& Hybrid RAG & 0.174 & 0.204 & 0.093 & 0.194 \\
& CoI-style & 0.168 & 0.188 & 0.085 & 0.188 \\
& ResearchAgent-style & 0.152 & 0.175 & 0.083 & 0.187 \\
& ARIS-style & 0.164 & 0.194 & 0.071 & 0.190 \\
\bottomrule
\end{tabular}
\end{table*}

Table~\ref{tab:error-low-counts} reports the selected low-score rates by family and method.
Planning has the largest reported low-Fact and low-FTA shares. Family differences also reflect
their distinct scoring functions, particularly semantic matching for Direction and Bottleneck
versus ranking alignment for Planning.

The Fact--FTA comparison finds 122 rows with an absolute score gap of at
least 0.5, alongside 20 high-FTA/low-Fact cases. These high-FTA/low-Fact cases are mostly in
Direction Forecasting, where the answer is near the future target space but misses the exact
mechanism, trajectory label, or scope. These cases illustrate differences between claim-level
validation and semantic target matching.

Planning diagnostics show that the dominant failure is top-priority error: within planning rows, the
top-priority-wrong rate ranges from 0.468 for ResearchAgent-style to 0.586 for Native LLM. These
answers often justify individual candidates well but do not support the full global ordering.
Venue diagnostics report pass rates of 0.877, 0.847, and 0.243 for the profile-evidence,
venue-bridge, and tradeoff-evidence links over evidence-grounded answers, respectively.
Tradeoff evidence is the least frequently supported of the three Venue links.

\begin{table*}[t]
\centering
\scriptsize
\setlength{\tabcolsep}{3.2pt}
\renewcommand{\arraystretch}{0.94}
\caption{Method-internal fingerprints from the diagnostic error audit. The table summarizes why
low-scoring cases fail within each method.}
\label{tab:method-internal-errors}
\begin{tabular}{p{0.16\linewidth}p{0.30\linewidth}p{0.44\linewidth}}
\toprule
Method & Fingerprint & Diagnostic interpretation \\
\midrule
Native LLM & Fluent but unanchored adjacent answers. &
Trace is not applicable because no external support artifact is exposed; factual and persuasive
errors often come from plausible priors rather than the target decision. \\
Hybrid RAG & Retrieved context without enough decision grounding. &
Retrieved evidence is often topical, but the final answer may not connect it to the required
bottleneck, ranking, or venue-conditioned comparison. \\
CoI-style & Strong local evidence, wrong global decision. &
The reasoning chain can be highly auditable while promoting a locally supported technical issue, such
as TOCTOU, into the root decision object. \\
ResearchAgent-style & Coherent scaffold, sometimes wrong root premise. &
Its coherent scaffold can still amplify a wrong early abstraction into a complete causal story. \\
ARIS-style & Specific mechanism over-commitment. &
Concrete mechanism selection helps Bottleneck tasks, but can narrow Direction forecasts toward a
specific instance when the reference target is a broader future mechanism cluster. \\
\bottomrule
\end{tabular}
\end{table*}

\subsection{Drift--Metric Associations}
\label{sec:appendix-bias-metric-coupling}

\begin{figure*}[t]
\centering
\includegraphics[width=\linewidth]{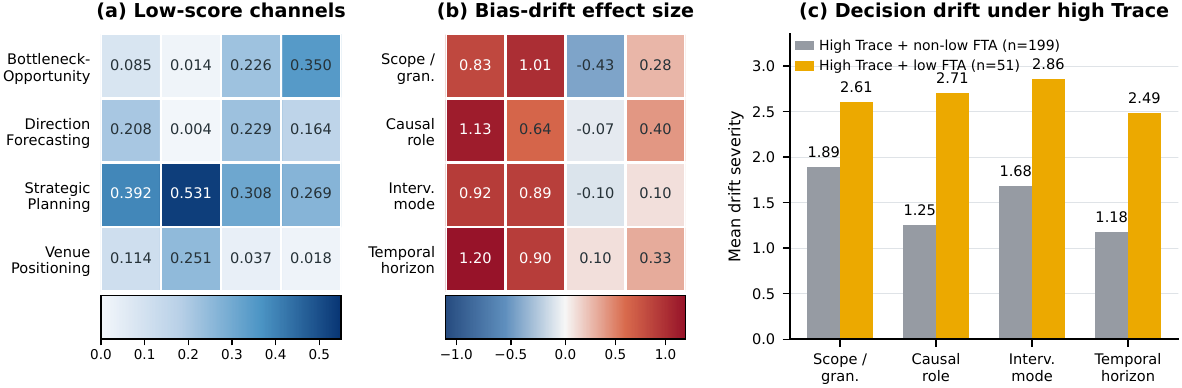}
\caption{Full diagnostic drift--metric audit. (a) Family-dependent low-score channels; (b) normalized
metric contrasts associated with the four drift dimensions; (c) drift severity among high-Trace answers.
The figure uses the complete sample of 960 hash-matched annotated answers.}
\label{fig:bias-metric-coupling-summary}
\end{figure*}

We join 960 drift-annotated answers with the formal metrics, matching all answers by hash.
Severity 0 denotes alignment and severity at least 2 denotes moderate-to-severe drift.
Table~\ref{tab:bias-metric-coupling} gives the numeric contrasts underlying
Figure~\ref{fig:bias-metric-coupling-summary}. Temporal-horizon annotations are retained in the
released analysis files and included in the normalized effect-size heatmaps.

\begin{table*}[t]
\centering
\scriptsize
\setlength{\tabcolsep}{3.5pt}
\caption{Diagnostic coupling between answer-reference drift dimensions and formal metric failures.
Metric drops are normalized by the standard deviation of the corresponding metric over all audited
rows, so they are expressed in standard-deviation units. All numbers come from the matched
drift-annotation sample joined with the formal metrics.}
\label{tab:bias-metric-coupling}
\begin{tabular}{p{0.18\linewidth}p{0.22\linewidth}p{0.19\linewidth}p{0.32\linewidth}}
\toprule
Question & Diagnostic contrast & Main result & Interpretation \\
\midrule
Causal-role drift vs. Fact &
Aligned vs. moderate-to-severe causal-role drift &
Fact drops by 1.11 SD under high causal-role drift &
Fact failures often arise when the answer assigns the wrong causal role to a plausible technical
object, such as treating a symptom as a root bottleneck or a solution as the forecast target. \\
Scope/granularity drift vs. FTA &
Aligned vs. moderate-to-severe scope drift &
FTA drops by 1.19 SD overall; Planning FTA drops by 1.94 SD &
The Planning contrast links scope mismatch to errors in the global ranking or priority structure. \\
Intervention-mode shift vs. FTA &
Mode-aligned vs. mode-shifted answers &
FTA drops by 1.10 SD overall; Planning FTA drops by 2.04 SD &
Changing the action type, such as benchmark-first vs. training-first, directly changes planning
and venue decisions even when local rationales remain plausible. \\
High traceability but low FTA &
High-trace/non-low-FTA rows vs. high-trace/low-FTA rows &
181 high-trace/non-low-FTA rows vs. 57 high-trace/low-FTA rows; causal severity
1.105 $\rightarrow$ 2.772; intervention-mode severity 1.530 $\rightarrow$ 2.860 &
High traceability does not guarantee target alignment: evidence can support a coherent but
misaligned decision object. \\
\bottomrule
\end{tabular}
\end{table*}

\begin{figure*}[t]
\centering
\begin{tcolorbox}[
  enhanced,
  width=0.97\linewidth,
  colback=white,
  colframe=black!45,
  boxrule=0.6pt,
  arc=1.5pt,
  left=5pt,
  right=5pt,
  top=5pt,
  bottom=5pt,
  fonttitle=\bfseries,
  title={High Traceability But High Drift: A Strategic-Planning Case}
]
\scriptsize
\begin{tabular}{p{0.20\linewidth}p{0.73\linewidth}}
\toprule
Case & Qwen3-235B + CoI-style, RFPLAN-0136, strategic research planning \\
\midrule
Task & Rank five visual world-model research directions for a six-month window, using only
evidence available by 2025-09-30. \\
CoI-style answer & Ranks \textbf{embodied evaluation} last because standardized metrics and tooling
remain immature, treating this gap as a reason to defer the direction. \\
Reference target & Ranks \textbf{embodied evaluation and feedback infrastructure} first because the
field already has plausible generative mechanisms but lacks evidence that they improve embodied decisions. \\
Scores & Trace = \textbf{0.892}, Fact = 0.343, FTA = \textbf{0.200}, Pers. = 0.591. \\
Drift diagnosis & The answer gives a traceable rationale for a plausible ordering but treats missing
evaluation infrastructure as a reason to deprioritize it rather than as the field's enabling bottleneck,
producing causal-role, scope, and intervention-priority drift. \\
\bottomrule
\end{tabular}
\end{tcolorbox}
\caption{A high-traceability/high-drift planning example, showing the task, agent decision,
reference target, and scores from the matched drift audit.}
\label{fig:high-trace-high-drift-case}
\end{figure*}

\begin{figure*}[t]
\centering
\includegraphics[width=\linewidth]{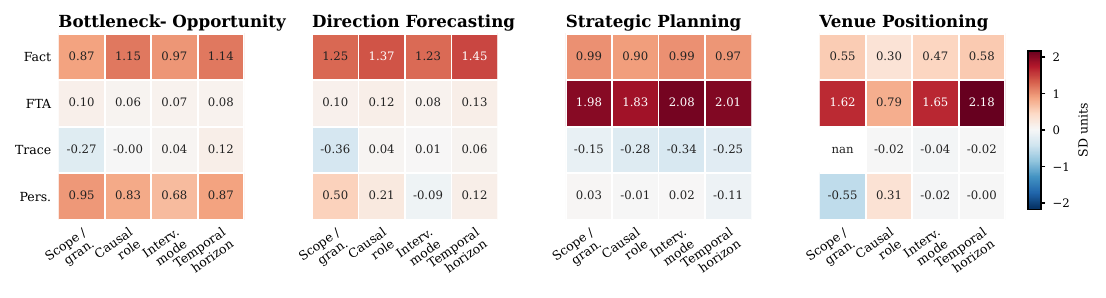}
\caption{Family-level drift--metric associations. Each heatmap reports normalized score
drop, $(\mathbb{E}[m\mid s=0]-\mathbb{E}[m\mid s\ge2])/\mathrm{SD}(m)$, for one task family. Rows are
formal metrics and columns are answer-reference drift dimensions.}
\label{fig:bias-metric-coupling-family-breakdown}
\end{figure*}

In RFDIR-0104, a GPT-5.2 ResearchAgent answer
has high Traceability (0.842), PERS 0.733, and moderate FTA (0.638), but zero
Prediction Factuality because it turns the target direction into a standardized PEFT evaluation
card, whereas the reference expects adaptive sparse zeroth-order or hybrid PEFT.
Figure~\ref{fig:high-trace-high-drift-case} gives a complementary Planning example in which the
answer and reference assign different investment priorities to the same evaluation gap.

\subsection{Task-specific Drift Mechanisms}
\label{sec:appendix-drift-mechanisms}

The drift audit identifies task-specific decision failures. In Bottleneck tasks, agents may promote a
symptom, local mechanism, or proposed solution to the system-level root cause. In Direction tasks,
single-paper evidence or a research suggestion may be generalized into a field trend, or new
retrieval results may displace a sound incumbent direction. In Planning tasks, evidence that a
method works can substitute for evidence that it is the best next research investment. Agents may
also compare a small implementation of one candidate with the cost of building another from
scratch. ARIS cases show commitment to a retrieved technical approach; ResearchAgent cases
include changes to previously aligned rankings. CoI can turn a locally coherent chain
around mature evidence into a global priority judgment even when maturity is not the relevant
investment criterion. In Venue tasks, agents may borrow the contribution of a related paper,
mistake topical overlap for contribution fit, or treat a proposed evidence upgrade as an established
property of the project. These substitutions can change both the default target and the conditions
under which an alternative becomes appropriate.

\paragraph{Bottleneck controls.}
We extract evidence-supported chains linking observed failures, proximal
mechanisms, and upstream causes. A task-level adjudicator checks explanatory coverage, abstraction,
and coupled causes, then locks the root bottleneck, causal bridge, and one-hop opportunity before
rendering. Chain extraction and rendering use Qwen3-235B; adjudication uses DeepSeek-v4-Pro with
reasoning enabled.

\paragraph{Direction controls.}
ARIS and ResearchAgent retain their incumbent direction among
the candidates considered during evidence retrieval and synthesis. CoI instead focuses retrieval
on its incumbent. The final decision is based on evidence matching the task's temporal and
conceptual scope.

\paragraph{Planning controls.}
Qwen first describes a feasible study, its decision value, and its uncertainty for
each public candidate using the task window and existing sources. A second stage reconciles these
studies with the original answer. ARIS commits only independently supported changes to the first-ranked
candidate. ResearchAgent diagnoses whether the current decision bottleneck is measurement,
capability, or integration; a second check commits only a supported change to the first-ranked candidate.
CoI first constructs an evidence-conditioned candidate ranking at a common project scale.
An independent gate then commits
a changed ranking only when the decisive comparison uses the correct evidence role and is supported
by an exact passage from the public task or supplied evidence. Maturity, frequency, and local
implementation details cannot alone justify a priority change. Each agent-specific rule was frozen
on 20 tasks before evaluation on the remaining 105.

\paragraph{Venue controls.}
Qwen constructs a \emph{public contribution contract} from the project description:
the contribution center, project kind, exact supporting passages, and current versus conditional
fit to the candidate venue families. A second stage checks the original answer against this contract.
Revision is constrained to contribution substitutions supported by exact task and answer passages:
related work presented as the project's own contribution, a future upgrade presented as current
evidence, or topical overlap used in place of contribution fit. The revised answer must retain a
complete positioning decision, including a default target, conditional alternatives, contribution
framing, reviewer risks, and evidence upgrades. If no supported conflict is found, or a revision
fails its passage or format checks, the original answer is retained. The contract and revision
rules were frozen on 20 development tasks
before the 105-task confirmation. Venue retrieval, tasks, and gold targets are held fixed.

\paragraph{Matched evaluation.}
Each family contains 125 tasks, split into 20 discovery/development tasks and 105 disjoint
confirmation tasks. Evaluation uses Qwen3-235B and the 105 confirmation tasks for all three agents:
315 baseline--intervention answer pairs and 1,260 paired dimension judgments per family.
Both arms include the complete public question and cutoff.
Generation uses the public question and cutoff-eligible evidence; hidden targets and drift scores
are evaluation-only. Failed revisions retain the original answer and remain in the denominator.
Qwen evaluates drift with the same
four-dimension protocol in both arms. We bootstrap tasks, keeping their agents and dimensions
together, using 10,000 resamples with seed 20260918 for Bottleneck and Direction and seed
20260915 for Planning. Table~\ref{tab:drift-mechanism-agent-effects} reports per-agent effects.
For Venue, both arms are judged separately with the full public question and current reference,
without agent identity, metric values, or revision rationale. Venue intervals use 10,000 task-bootstrap
resamples with seed 20260918, keeping all three agents together within each sampled task.

\begin{table}[htbp]
\centering
\small
\setlength{\tabcolsep}{5pt}
\caption{Per-agent drift effects on 105 disjoint confirmation tasks per family (420 paired dimension
judgments per row), excluding the 20 discovery/development tasks. Lower is better.
Planning uses separately frozen controls for each agent. RA denotes ResearchAgent.}
\label{tab:drift-mechanism-agent-effects}
\begin{tabular}{llrrr}
\toprule
Task & Agent & Before & After & $\Delta$ [95\% CI] \\
\midrule
Bottleneck & ARIS & 1.600 & 1.352 & $-0.248$ [$-0.381$, $-0.119$] \\
 & CoI & 1.633 & 1.290 & $-0.343$ [$-0.488$, $-0.195$] \\
 & RA & 1.576 & 1.333 & $-0.243$ [$-0.376$, $-0.112$] \\
\midrule
Direction & ARIS & 1.793 & 1.679 & $-0.114$ [$-0.198$, $-0.038$] \\
 & CoI & 1.569 & 1.562 & $-0.007$ [$-0.071$, $0.052$] \\
 & RA & 1.610 & 1.550 & $-0.060$ [$-0.126$, $0.002$] \\
\midrule
Planning & ARIS & 1.738 & 1.519 & $-0.219$ [$-0.343$, $-0.107$] \\
 & CoI & 1.557 & 1.502 & $-0.055$ [$-0.207$, $0.088$] \\
 & RA & 1.371 & 1.279 & $-0.093$ [$-0.238$, $0.033$] \\
\bottomrule
\end{tabular}
\end{table}

On the 105-task confirmation set, ARIS commits 18 revisions and changes drift/FTA from
1.7381/0.5980 to 1.5190/0.6681, while ResearchAgent commits 14 revisions and changes them from
1.3714/0.6870 to 1.2786/0.7049. CoI commits 27 revisions and changes drift/FTA from
1.5571/0.6440 to 1.5024/0.6640. FTA increases for all three Planning agents.
The individual CoI and ResearchAgent drift confidence intervals include zero; the combined
Planning drift interval excludes zero. Planning rank alignment maps equivalent
candidate names to their public option identities before scoring.

\paragraph{Content verification.}
For Bottleneck, Direction, and Planning, content verification uses the 105 confirmation tasks
per family and all three agents. Fact uses DeepSeek-v4-Pro in both arms, while FTA uses the
family-specific semantic or ranking formulation. Unchanged answers reuse
their exact matched results. Table~\ref{tab:drift-mechanism-content-effects} presents per-agent
content scores. The aggregate confirmation effects are summarized in the main text
(Figure~\ref{fig:drift-intervention-summary}). Bottleneck gains 0.089 in Fact and 0.005 in FTA;
Direction Fact increases from 0.562 to 0.579 and FTA from 0.632 to 0.651;
Planning gains 0.066 in Fact and 0.036 in FTA. FTA increases in all four task families.

\begin{table}[htbp]
\centering
\small
\setlength{\tabcolsep}{6pt}
\caption{Matched content verification with Qwen3-235B on 105 disjoint confirmation tasks per
family, excluding the 20 discovery/development tasks. All three agents are evaluated; displayed rows present
per-agent Fact and FTA scores, both higher-is-better. RA denotes ResearchAgent.}
\label{tab:drift-mechanism-content-effects}
\begin{tabular}{llrrrr}
\toprule
 & & \multicolumn{2}{c}{Fact} & \multicolumn{2}{c}{FTA} \\
\cmidrule(lr){3-4}\cmidrule(lr){5-6}
Task & Agent & Before & After & Before & After \\
\midrule
Bottleneck & ARIS & 0.8600 & 0.9242 & 0.6070 & 0.6131 \\
 & CoI & 0.8510 & 0.9555 & 0.6070 & 0.6117 \\
 & RA & 0.8020 & 0.9005 & 0.6080 & 0.6116 \\
\midrule
Direction & ARIS & 0.5420 & 0.5590 & 0.6270 & 0.6440 \\
 & CoI & 0.5770 & 0.5950 & 0.6350 & 0.6520 \\
 & RA & 0.5680 & 0.5840 & 0.6330 & 0.6570 \\
\midrule
Planning & ARIS & 0.4600 & 0.5245 & 0.5980 & 0.6681 \\
 & CoI & 0.5040 & 0.5700 & 0.6440 & 0.6640 \\
 & RA & 0.5320 & 0.5980 & 0.6870 & 0.7049 \\
\bottomrule
\end{tabular}
\end{table}

\paragraph{Venue confirmation.}
Of the 315 Venue pairs, 197 answers are revised and 118 retain the original, including 19 failed
revision attempts; all remain in the evaluation denominator. Fact uses claim-level F1 over
answer-claim support and hidden claim-bank coverage in both arms, with Qwen judging the paired
answers against the same fixed hidden claim bank.
FTA retains the benchmark's default-family match, conditional-alternative recall, and contribution-center
alignment, using identical canonicalization and a shared embedding cache. Unchanged answers reuse
their exact matched scores. Overall Fact increases from 0.5119 to 0.5450
($\Delta=0.0331$, 95\% CI $[0.0125,0.0539]$), and FTA from 0.6916 to 0.7437
($\Delta=0.0521$, 95\% CI $[0.0044,0.0972]$). Mean drift falls from 1.4778 to 1.2944
($\Delta=-0.1833$, 95\% CI $[-0.2595,-0.1056]$).
At the task level, changes in scope and intervention-mode drift are negatively associated with
FTA changes (Spearman $\rho=-0.366$ and $-0.686$, respectively), whereas causal-role changes have
little association with Fact changes ($\rho=-0.057$). These are descriptive associations across
the paired task means, not causal estimates of individual drift dimensions.

\FloatBarrier

\end{document}